\documentclass[10pt,twocolumn,letterpaper]{article}

\usepackage[pagenumbers]{cvpr} 
\definecolor{cvprblue}{rgb}{0.21,0.49,0.74}
\usepackage[pagebackref,breaklinks,colorlinks,allcolors=cvprblue]{hyperref}
\usepackage{graphicx}
\usepackage{multirow}

\def\paperID{15656} 
\def\confName{CVPR}
\def\confYear{2025}

\title{CAD-Llama: Leveraging Large Language Models for Computer-Aided Design Parametric 3D Model Generation}

\author{
    Jiahao Li\quad Weijian Ma\quad Xueyang Li\quad Yunzhong Lou\quad Guichun Zhou\quad Xiangdong Zhou\thanks{Corresponding author.}\\
    School of Computer Science and Technology, Fudan University \\
    {\tt\small \{lijh23, mawj22, xueyangli21\}@m.fudan.edu.cn} \quad
    {\tt\small \{yzlou20, gczhou19, xdzhou\}@fudan.edu.cn}
}

\begin{document}
\twocolumn[{
\renewcommand\twocolumn[1][]{#1}
\maketitle
\begin{center}
        \fontsize{10.25pt}{\baselineskip}\selectfont
\end{center}
\vspace{.4mm}
    \begin{center}
        \vspace{-20mm}
        \centering
        \includegraphics[width=1\linewidth]{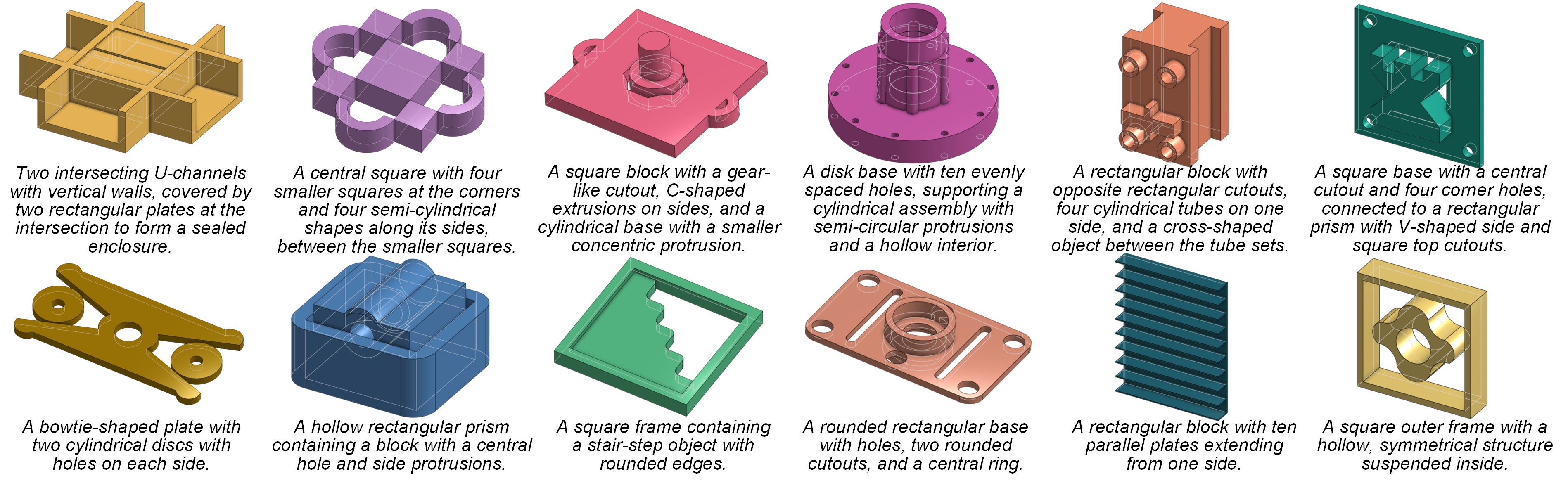}
        \vspace{-6mm}
        \captionof{figure}{\textbf{A collection of generated CAD models with text prompts by using our method (CAD-Llama-INS).} Our approach enables the generation of more complex CAD models based on both abstract and detailed text prompts.}
        \label{fig:figure1}
    \end{center}
}]
{\let\thefootnote\relax\footnotetext{* Corresponding author.}}
\begin{abstract}
  Recently, Large Language Models (LLMs) have achieved significant success, prompting increased interest in expanding their generative capabilities beyond general text into domain-specific areas. This study investigates the generation of parametric sequences for computer-aided design (CAD) models using LLMs. This endeavor represents an initial step towards creating parametric 3D shapes with LLMs, as CAD model parameters directly correlate with shapes in three-dimensional space. Despite the formidable generative capacities of LLMs, this task remains challenging, as these models neither encounter parametric sequences during their pretraining phase nor possess direct awareness of 3D structures. To address this, we present CAD-Llama, a framework designed to enhance pretrained LLMs for generating parametric 3D CAD models. Specifically, we develop a hierarchical annotation pipeline and a code-like format to translate parametric 3D CAD command sequences into Structured Parametric CAD Code (SPCC), incorporating hierarchical semantic descriptions. Furthermore, we propose an adaptive pretraining approach utilizing SPCC, followed by an instruction tuning process aligned with CAD-specific guidelines. This methodology aims to equip LLMs with the spatial knowledge inherent in parametric sequences. Experimental results demonstrate that our framework significantly outperforms prior autoregressive methods and existing LLM baselines.
\end{abstract}
\vspace{-6mm}
\section{Introduction}
\label{sec:intro}
 Computer-Aided Design (CAD) generative modeling has attracted increasing attention from research and industry communities.
Large language models (LLMs) have recently demonstrated strong
generative capabilities and impressive zero-shot performance across a
broad range of downstream tasks  \cite{luo2023wizardmath, yu2023metamath, yue2023mammoth, collins2023evaluating}. These models have also
found widespread applications in the real world \cite{madani2023large, ferruz2022controllable, bubeck2023large, liu2024conversational}.
However, the exploration of utilizing LLMs for generating parametric CAD construction sequences remains underexplored, thereby calling for further investigation
on how to invoke the potential of LLM's learned priors onto the task of parametric CAD sequence generation and editing.

Leveraging LLMs for parametric CAD sequence generation is nontrivial. A substantial disparity exists between the original parameterized CAD sequences and the natural language familiar to LLMs, rendering the direct generation of parametric CAD sequences by LLMs a challenging task. Most previous works reconstruct parametric CAD sequences from various inputs, such as point clouds \cite{Ma2024DrawStep}, text \cite{wu2021deepcad, khan2024text2cad, Li2024CADTranslator}, B-rep models \cite{willis2021engineering, xu2021inferring}, and partial CAD \cite{xu2022skexgen, xu2023hierarchical}, using encoder-decoder architectures trained solely on CAD dataset \cite{wu2021deepcad, khan2024text2cad, Li2024CADTranslator}. Some recent attempts demonstrate that LLMs can generate basic CAD construction sequences \cite{Wu2023CADLLM, Li2024LLM4CAD, Yuan2024OpenECAD, Badagabettu2024Query2CAD} and have the potential to understand the semantics of symbolic graphic programs \cite{Qiu2024LLM_SGP}. However, most of these methods suffer from weak generalization and lack the ability to generate complex CAD models, let alone generate CAD models based on complex text instructions.

We note that in order to effectively employing the generative capabilities of LLMs for CAD sequence generation necessitates a comprehensive understanding of both the characteristics of CAD data and the intrinsic capabilities of LLMs. The parametric CAD model, also referred to as CAD design history, consists of sequences of commands from CAD tools, yet it lacks semantic annotations pertaining to the design rationale and the geometry or shape of the respective CAD model. Consequently, without textual descriptions, it is challenging for LLMs to grasp the semantic implications of parametric CAD models. This limitation accounts for the fact that, in prior research, LLMs have typically only generated relatively simple CAD models. Conversely, LLMs excel in code generation owing to the extensive repository of code data accompanied by text comments and functional descriptions present in the training datasets. 

Leveraging insights from the CAD modeling process and the remarkable language generation capabilities of LLMs, we propose CAD-Llama, an extensive framework that adapts open-source LLMs for the generation of CAD command sequences. For data acquisition, we introduce a novel hierarchical data annotation pipeline for CAD design history data, which is represented in the form of Python-like code, called \textbf{S}tructured \textbf{P}arametric \textbf{C}AD \textbf{C}ode (SPCC). During the annotation process, a visual language model (VLM) is utilized to annotate both the three-dimensional geometry and the two-dimensional sketch of each component with detailed textual descriptions. Subsequently, the comprehensive semantics and the interrelationships among components are captioned to yield the top-layer thorough descriptions. Regarding training methodologies, an adaptive pretraining paradigm, in conjunction with instruction tuning techniques for varied downstream tasks, is proposed to impart CAD modeling capabilities to the LLM and to adapt it for diverse downstream applications.

We conduct a series of experiments to evaluate our
approach, covering both unconditional and conditional generation tasks.
The results indicate that our method outperforms recent state-of-the-art
parametric CAD generation models, as well as open-source models such as
GPT-4 and LLaMA3, across various CAD-related tasks. We show that using rich and structured text descriptions of 3D shape and geometry to fine-tune LLMs leads to the emergence of the ability to generate professional parametric 3D CAD models under complex text instructions, as shown in Figure \ref{fig:figure1}, which has not been explored or reported in previous studies.

In summary, our contributions include the following.

\begin{enumerate}
\def\labelenumi{\arabic{enumi}.}
\item
We present  CAD-Llama, a novel unified framework to leverage LLMs' generative priors for parametric 3D CAD modeling based on text instructions.
\item
We introduce a hierarchical annotation pipeline for 3D CAD models that
captures both structured information and detailed textual descriptions of 3D shapes and geometry.
\item 
We propose an adaptive pretraining paradigm combined with instruction tuning on a multitask instructional dataset to align LLMs with CAD sequence modeling across a series of tasks.
\item
Experimental results demonstrate that CAD-Llama can generate more accurate and complex parametric CAD models and achieve good performance in a series of downstream tasks.
\end{enumerate}

\section{Related Work}
\label{sec:related work}
\noindent\textbf{Representation Learning of CAD models.}
Building representations for understanding CAD models has become a long-sought problem throughout the vision history. Early research focused on utilizing shape-understanding methods to classify and segment CAD models in form of point clouds \cite{qi2017pointnet, qi2017pointnet++}, meshes \cite{spacemesh2024, gao2023learning}, voxels \cite{Maturana2015VoxNet, Riegler2017OctNet, Liu2019PVCNN} and SDFs \cite{Park_2019_CVPR, Chabra2020DeepLS}.
However, methods in the shape domains fail to capture the exact shape parameters, leading to a difficulty in editing and reusing the created shapes.  On the other hand, along with the emergence of large-scale parametric CAD datasets \cite{wu2021deepcad, koch2019abc, willis2021fusion}, language models have been adopted to model the parametric designs of CAD models. \cite{Ma2023MultiCAD} also built a multimodal representation for CAD models based on point clouds and construction sequences. The sequence modeling ability of language models has opened up possibilities of generating precise parametric construction sequences. However, the granularity of control over parameters still remains a problem.

\begin{figure*}[!h]
    \centering
    \vspace{-6mm}
	\includegraphics[width=\linewidth]{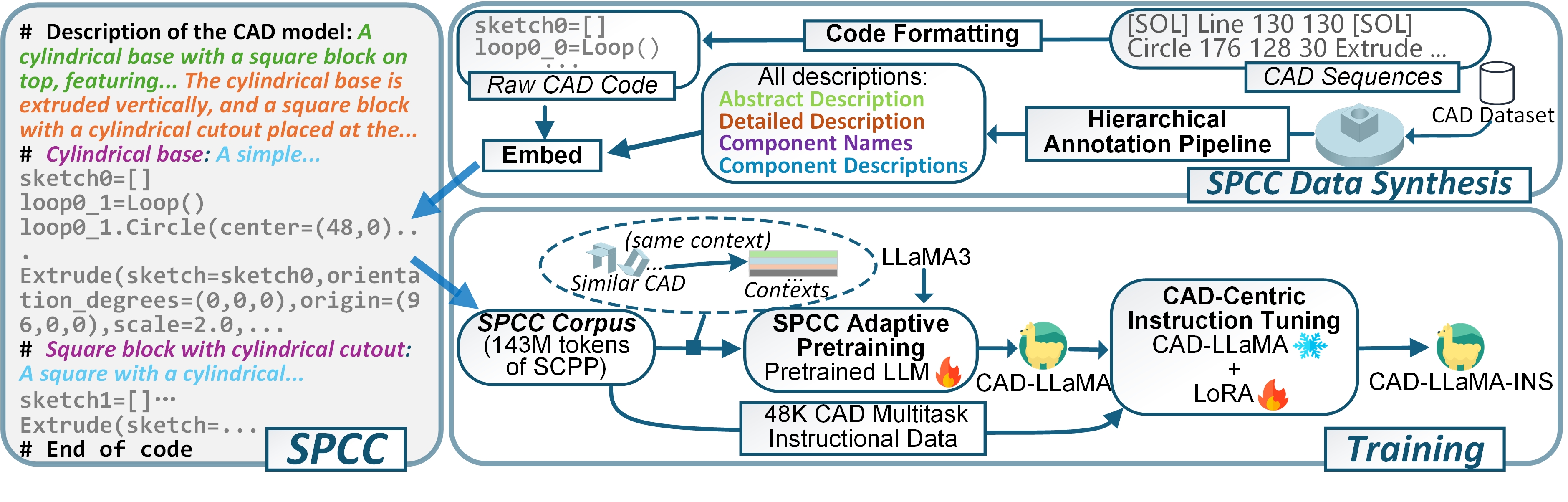}
    \vspace{-8mm}
	\caption{\textbf{Overview of the proposed framework CAD-Llama.} The framework consists of two parts: (1) SPCC data synthesis, which employs the hierarchical annotation pipeline to convert CAD sequences into SPCC representations, and (2) the pretraining and instruction tuning process, where the resulting SPCC corpus is leveraged to enhance model performance.}
    \vspace{-4mm}
 \label{fig:input_engineering}
\end{figure*}

\noindent\textbf{Crossmodal CAD Generation.}
Translating parametric CAD models from given conditions has been a problem of active research. Research in earlier times focused on precise translation from geometric shapes like point clouds or meshes into parametric sequences via heuristic primitive fitting methods like RANSAC \cite{fischler1981random, schnabel2007efficient} or Hough Transform \cite{duda1972use, borrmann20113d}. Some follow-up works attempted to broaden the scope of input modalities. They are Point cloud\cite{wu2021deepcad, Ma2024DrawStep}, partial CAD input \cite{xu2023hierarchical}, target B-reps \cite{willis2021engineering, xu2021inferring}, voxel grids\cite{li2023secad, lambourne2022reconstructing},  point clouds with\cite{uy2022point2cyl} or without sequence guidance \cite{ren2022extrudenet, li2023secad}, etc. However, all these works require detailed semantics of the target models, limiting their applications to the domains of concept design. Concurrent works on CAD model generation from text descriptions include Text2CAD \cite{khan2024text2cad} and CAD Translator \cite{Li2024CADTranslator}, both of which employ encoder-decoder architectures to translate the text descriptions of CAD shapes into parametric CAD sequences. However, the limited capacity of the encoder-decoder architecture constrains its generalizability on out-of-distribution examples.

\noindent\textbf{Large Language Models and Computer-Aided Design.}
LLMs have demonstrated growing potential in many applications, ranging from mathematical problem solving and theorem proving assistance \cite{luo2023wizardmath, yu2023metamath, yue2023mammoth, collins2023evaluating} to aiding
biological discovery \cite{madani2023large, ferruz2022controllable, bubeck2023large, liu2024conversational}. Applying LLMs for abstract content understanding and generation is also a popular direction of research. 
A recent investigation \cite{Qiu2024LLM_SGP} shows that LLMs can understand symbolic graphic programs like SVG and CAD models via finetuning on VQA tasks. A few attempts tried to investigate the generation ability of LLMs on parametric CAD models. CAD-LLM \cite{Wu2023CADLLM} empirically investigates CAD generation on 2D domains. LLM4CAD \cite{Li2024LLM4CAD} utilizes VLMs to perform zero-shot CAD generation tasks. CADTalk \cite{yuan2024cadtalk} generates semantic labels for CAD parts. OpenECAD \cite{Yuan2024OpenECAD} attempts to finetune a VLM with the assistance of CAD kernels like PythonOCC. Query2CAD \cite{Badagabettu2024Query2CAD} proposes a natural language translator into CAD code with an image-captioner in the loop. CAD-MLLM \cite{Cad-mllm} and CAD-GPT \cite{CAD-GPT} both leverage Multimodal Large Language Models (MLLMs) that generate CAD command sequences, with CAD-MLLM supporting diverse inputs like text, images, and point clouds, and CAD-GPT enhancing spatial reasoning for precise synthesis from single-view images or text. However, few previous work succeeded in leveraging LLM's strong generative prior on text to CAD construction sequence generation.

\vspace{-2mm}
\section{Method}
In this section, we first propose the hierarchical annotation pipeline and the SPCC dataset synthesis for LLMs finetuning data preparation. Then, we propose a pretraining method to equip LLMs with CAD model generation capabilities, and an instruction tuning method that further leverage the LLM's ability to handle CAD-related downstream tasks.  The framework of CAD-Llama is illustrated in Figure \ref{fig:input_engineering}. Details are provided in the following subsections.

\begin{table}[t]
\centering
\resizebox{\linewidth}{!}{
\begin{tabular}{c|c}
\toprule
\textbf{Notation} & \textbf{Definition} \\
\midrule
\({\mathcal{D}}_{j}\) & \parbox[c]{7.1cm}{\centering The \textit{i}-th CAD in dataset} \\
\midrule
\({\mathcal{D}}_{j\setminus {\mathcal{P}_j^k}}\) & \parbox[c]{7.1cm}{\centering CAD of \({\mathcal{D}}_{j}\) after removing \textit{k}-th component } \\
\midrule
\(C(\mathcal{D}_j)\) & \parbox[c]{7.1cm}{\centering CAD code representation of \({\mathcal{D}}_{j}\)} \\
\midrule
\(\tilde{\mathcal{D}_j}\) & \parbox[c]{7.1cm}{\centering SPCC representation of \({\mathcal{D}}_{j}\)} \\
\midrule
\(\mathcal{A}_j,\mathcal{T}_j\) & \parbox[c]{7.1cm}{\centering Overall abstract and detailed descriptions of \({\mathcal{D}_{j}}\) } \\
\midrule
\(\mathcal{S}_j^i,\mathcal{I}_{j}^i\) & \parbox[c]{7.1cm}{\centering Name and description of the \textit{i}-th component of 
\({\mathcal{D}_{j}}\)} \\
\midrule
\end{tabular}
}
\vspace{-3mm}
\caption{Summary of key notations.}
\label{tab:key_notations}
\vspace{-4mm}
\end{table}

\subsection{Notation}\label{notation}
Denote the data set of the parametric CAD model as
\(\mathcal{D}=\{\mathcal{D}_1,\mathcal{D}_2,\dots,\mathcal{D}_N\}\),
where \(N\) is the number of CAD models. Assume that
\(j\)-th CAD model \(\mathcal{D}_j\) contains \(m\) components,
represented as
\(\mathcal{D}_j=\{\mathcal{P}_j^1,\mathcal{P}_j^2,\dots,\mathcal{P}_j^m\}\),
where \(\mathcal{P}_j^i\) refers to the parametric CAD sequence of the
\(i\)-th component. 
In Table \ref{tab:key_notations}, we provide a brief description of the key notation. A detailed introduction to these notation is presented in the following two subsections.

\begin{figure*}[!h]
    \centering
    \vspace{-6mm}
	\includegraphics[width=\linewidth]{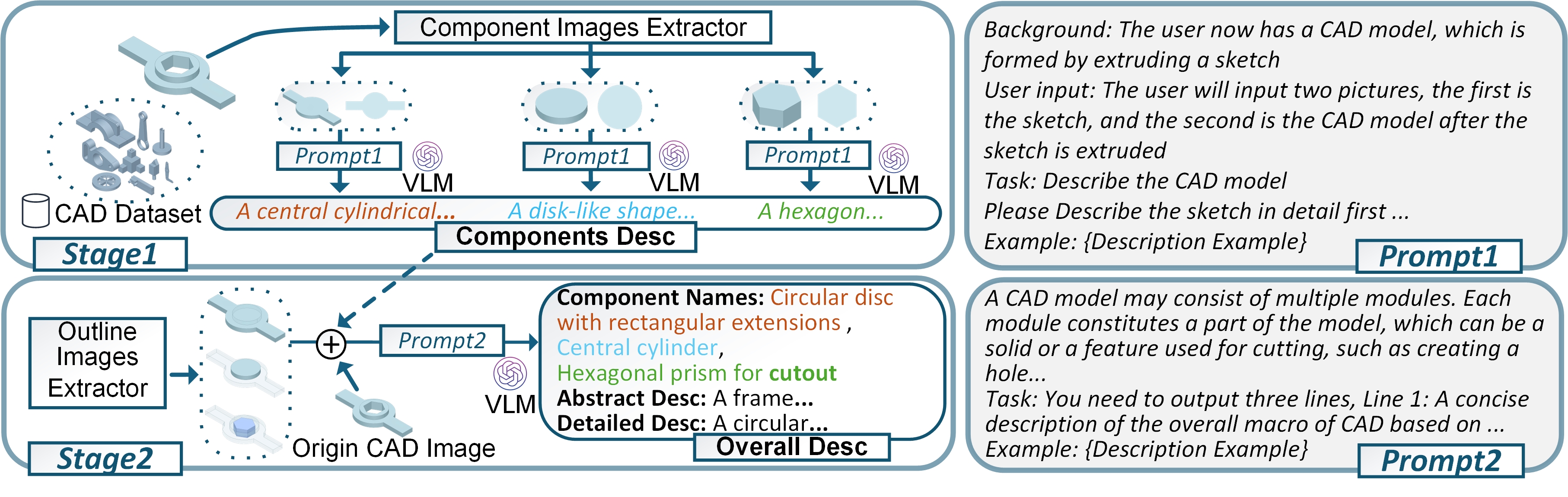}
    \vspace{-8mm}
	\caption{\textbf{Hierarchical Annotation Pipeline}. The figure illustrates our two-stage annotation process for CAD models. In the first stage, detailed descriptions of individual components are generated. In the second stage, a global description is produced, which includes both an abstract overview and detailed descriptions that capture the spatial relationships between components.}
    \vspace{-4mm}
 \label{fig:annotation_pipeline}
\end{figure*}

\subsection{Hierarchical Annotation Pipeline}\label{sec:annotation-pipeline}

A crucial step in fine-tuning or training a domain-specific LLM is constructing a domain dataset that bridges the gap between plain language which LLMs understand well, and domain-specific data. For parametric CAD model generation, this involves annotating 3D CAD models with text descriptions. Prior work has utilized VLMs to generate simple text labels or brief descriptions for training datasets. However, we believe that more detailed, structured textual descriptions of 3D shapes are essential for effective LLM fine-tuning, an aspect underexplored in previous studies.

Using VLMs for comprehensive CAD model annotations presents challenges, as a single prompt often fails to capture both fine-grained geometric properties and compositional relationships. To address this, we propose a two-stage hierarchical annotation approach, as illustrated in Figure \ref{fig:annotation_pipeline}.

\textbf{Component Description Annotation} The first stage focuses on the detailed description of individual components. 
Formally, for the \( i \)-th component \( \mathcal{P}_j^i \) of \( \mathcal{D}_j \), we first produce the 3D image \( {I}_j^i \) (obtained by projecting the 3D model into the 2D image) and the 2D sketch image \( {\hat I}_j^i \) of \( \mathcal{P}_j^i \) (obtained by rendering the corresponding 2D sketch commands). We then feed these images into VLMs (GPT-4o \cite{achiam2023gpt} used in our experiment), generating a detailed description \( \mathcal{I}_{j}^i \) of the \( i \)-th component based on a pre-designed prompt:
\begin{equation}
\mathcal{I}_{j}^i=\text{VLM}(\text{prompt}_1,{I}_j^i,{\hat I}_j^i),
\end{equation}
where \( \text{prompt}_1 \) is the pre-designed prompt used in stage one. By applying the above process to each component, we obtain detailed descriptions of all components \( \mathcal{I}_j=\{\mathcal{I}_{j}^1,\mathcal{I}_{j}^2,\dots,\mathcal{I}_{j}^m\} \).
Additionally, we also include additional parameter information 
in the prompt, such as the extrusion
direction and extrusion length. Taking component 1 in Figure \ref{fig:annotation_pipeline} as an
example, the generated description is: "\emph{A central cylindrical
disk, with two symmetrically positioned rectangular blocks extending
from opposite sides of the disk\textquotesingle s circumference, forming
a shape that resembles a circular center with bar-like extensions,
extruded upwards with an extrusion length of 12 units}".

\textbf{Overall Description Annotation} The second stage focuses on global descriptions, which include an abstract overview as well as a detailed description that explicitly addresses the spatial relationships and assembly process of the components. Additionally, since the global and local descriptions are obtained in different stages, there may be some semantic discontinuity. To bridge this gap, we let the VLM (GPT-4o) generate a short name for each component to link global and local descriptions. 
Specifically, for \textit{m} components in $\mathcal{D}_j$, we first generate its outline images ${\dot I}_j=\{{\dot I}_j^1,{\dot I}_j^2,\dots,{\dot I}_j^m\}$ by enhancing the visibility of the target component and increasing the transparency of other components to clearly emphasize its specific location within the CAD model. Components used for \textit{Cutting} are rendered in blue. We then input these outline images ${\dot I}_j$, the original CAD image ${{I}}_j$, the descriptions for each module obtained in the first stage $\mathcal{I}_j$, and the $\text{prompt}_2$ used in the second stage into the VLM (GPT-4o) to generate the desired descriptions:
\begin{equation}
\mathcal{A}_j,\mathcal{T}_j,\mathcal{S}_j=\text{VLM}(\text{prompt}_2,{I}_j,\ \mathcal{I}_j,\ {\dot I}_j),
\end{equation}
where $\mathcal{A}_j$ and $\mathcal{T}_j$ are the overall abstract and detailed descriptions, respectively, and $\mathcal{S}_j=\{\mathcal{S}_{j}^1,\mathcal{S}_{j}^2,\dots,\mathcal{S}_{j}^m\}$ represents the short names for each component. For CAD models with a single component, the first-stage description serves as the final description.

These local and global hierarchical descriptions can be seamlessly integrated with the CAD data, which is designed similarly with a hierarchical structure, as detailed in Section~\ref{sec:spcc-sec}.
To enhance the stability and adaptability of VLM outputs to varying CAD model complexities, we classify CAD sequences into five complexity levels based on their length, providing 50 high-quality examples per level. All prompts employ a two-shot approach \cite{brown2020language}, selecting two examples from the corresponding level based on the complexity of the CAD model. This strategy reduce hallucinations \cite{liu2023aligning, sun2023aligning, jiang2024hallucination} and improve the overall output quality.
\subsection{SPCC Data Synthesis}\label{sec:spcc-sec}
Inspired by the considerable capabilities of LLMs in code generation \cite{gao2023makes, wong2023natural, ma2023training}, as well as
some studies \cite{wang2024instructgraph, Yuan2024OpenECAD} have attempted to convert various data types into a unified code format, we first convert parametric CAD sequences into a unified code format, as illustrated in the left part of Figure \ref{fig:input_engineering}.
Similarly to \cite{Yuan2024OpenECAD},
we represent each sketch as a list of loops (e.g., \texttt{sketch\_i.append(loop1)}),
where each loop can call methods like \textit{Line}, \textit{Arc}, or \textit{Circle} to draw. (e.g.,\texttt{loop1.Arc(endpoint=(87,-8),degrees=\allowbreak134,counterclockwise=True)})
Finally, the extrusion is performed by referencing the corresponding sketch to complete the operation.  
For the continuous parameters of the coordinates, we use the original 8-bit quantized parameters from \cite{wu2021deepcad}, where the starting point is set to (128, 128). To provide a more intuitive representation of scale information, we recenter the starting point to (0, 0). For angular parameters, we use discrete angle values within the range of 0 to 360 degrees. For more details, please refer to the supplementary materials.

\textbf{SPCC Corpus} Let \(C(·)\) denote the CAD code formatting process and $F(·)$ represent our 
proposed annotation pipeline. For the CAD model \(\mathcal{D}_j\), we obtain the parametric CAD code
representation \(C(\mathcal{D}_j)=\{C(\mathcal{P}_j^1),C(\mathcal{P}_j^2),\dots,C(\mathcal{P}_j^m)\}\) and all necessary annotations
\(F(\mathcal{D}_j)=\{\mathcal{I}_j,\mathcal{A}_j,\mathcal{T}_j,\mathcal{S}_j\}\), where $\mathcal{I}_j=\{\mathcal{I}_{j}^1,\mathcal{I}_{j}^2,\dots,\mathcal{I}_{j}^m\}$.
Next, we integrate annotations by embedding them into specific segments of the CAD code, creating the SPCC. 
Specifically, for the \(i\)-th component
\(\mathcal{P}_j^i\) of \(\mathcal{D}_j\), we combine its corresponding
code and annotations to get its final representation: 
\(\tilde{\mathcal{P}}_j^i=\{\text{concat}\{\mathcal{S}_{j}^i,\mathcal{I}_{j}^i\}, C(\mathcal{P}_j^i)\}\),
where concat represents the concatenate operation. This process produces each component's final representation. We then
add global descriptions as a prefix to obtain the final SPCC
representation of \(\mathcal{D}_j\), denoted as
\(\tilde{\mathcal{D}_j}=\{\mathcal{A}_j,\mathcal{T}_j,\tilde{\mathcal{P}}_j^1,\tilde{\mathcal{P}}_j^2,\dots,\tilde{\mathcal{P}}_j^m\}\),
resulting in the corpus
\(\tilde{\mathcal{D}}=\{\tilde{\mathcal{D}}_1,\tilde{\mathcal{D}}_2,\dots,\tilde{\mathcal{D}}_N\}\)
for training. Additionally, to enable LLMs to generate diverse CAD models from both detailed and abstract descriptions, we include data that contain only abstract descriptions,
denoted as
\(\dot{\mathcal{D}_j}=\{\mathcal{A}_j,\tilde{\mathcal{P}}_j^1,\tilde{\mathcal{P}}_j^2,\dots,\tilde{\mathcal{P}}_j^m\}\),
in the final training corpus, represented as
\(\dot{\mathcal{D}}=\{\dot{\mathcal{D}}_1,\dot{\mathcal{D}}_2,\dots,\dot{\mathcal{D}}_N\}\).
For models with only one component, such as \(\mathcal{D}_k\), we have
\(\tilde{\mathcal{D}}_k=\ \dot{\mathcal{D}_k}=\{\mathcal{I}_{k}^1, C(\mathcal{P}_k^1)\}\).
Thus, the final SPCC corpus is
\(\mathcal{D}_{\text{SPCC}}=\{\tilde{\mathcal{D}},\ \dot{\mathcal{D}}\}\).

\textbf{Multitask Instructional Dataset}
To adapt CAD-Llama for downstream tasks, we construct a suite of  CAD-centric instructional datasets, including \textit{text-to-CAD}, \textit{completion}, \textit{caption} (CAD description generation), \textit{addition}, and \textit{deletion}.
Table \ref{tab:graph_tasks} presents detailed information about each task, including task descriptions, inputs, and outputs.
Figure \ref{fig:editing_flow_examples} also provides an example of CAD-related tasks, demonstrating how this series aids designers in continuously optimizing the model, from initial construction to iterative refinement.
For \textit{completion}, 
we use the initial portion (approximately 30\% to 50\% in our experiments) of \(\tilde{\mathcal{D}_j}\) as input. 
For CAD editing tasks (\textit{addition} and \textit{deletion}),
not all operations are logically valid. For example, in a CAD model consisting of a solid component and a cutting component, deleting the solid component while retaining only the cutting component is illogical.
To effectively construct CAD editing instruction data using SPCC, we employ GPT-4o to identify the removable component \(k\) within \(\mathcal{D}_j\), explicitly justify the logical validity of its deletion, and generate corresponding deletion and inverse-addition instructions.
We then remove module \(k\) from \(\mathcal{D}_j\), obtaining the edited CAD model \({\mathcal{D}}_{j\setminus {\mathcal{P}_j^k}} \).
Using both \(\mathcal{D}_j\) and \({\mathcal{D}}_{j\setminus {\mathcal{P}_j^k}} \)
along with the instructions, we construct the dataset for \textit{addition} and \textit{deletion}. 

For the \textit{addition} and \textit{deletion} tasks, both input and output CAD representations are provided as CAD code, which lacks hierarchical descriptions.
To demonstrate that SPCC 
enhances CAD editing performance and 
that CAD-Llama effectively understands the inherent structure of CAD Code, we designed two variant tasks: \textit{deletion\textsuperscript{*}} and \textit{addition\textsuperscript{*}}.
During training, both input and output CAD models are ground-truth SPCC. During testing, we first use CAD-Llama-INS (instruction-tuned version of CAD-Llama) to caption the input CAD code, and the resulting SPCC serves as the final input CAD model.
Taking \textit{deletion\textsuperscript{*}} as an example, inputs and outputs at different stages are as follows:
\begin{align*}
\text{(Train)}\ &\quad\quad\quad\quad\text{Input: }\tilde{\mathcal{D}}_{j} \rightarrow\text{Output: }\tilde{\mathcal{D}}_{j \setminus P_k} \\
\text{(Test)}\ \ &\text{Input: }C(\mathcal{D}_j) \xrightarrow[\text{CAD-Llama-INS}]{\textbf{Caption}} \tilde{\mathcal{D}}_{j} \rightarrow \text{Output: }\tilde{\mathcal{D}}_{j \setminus P_k}
\end{align*}

\begin{table}[t]
\centering
\setlength{\tabcolsep}{1pt}
\resizebox{\linewidth}{!}{
\begin{tabular}{c|c|c|c}
\toprule
\textbf{Task Name} & \textbf{Definition} & \textbf{Input} & \textbf{Output} \\
\midrule
\textit{Text-to-CAD} & \parbox[c]{7.6cm}{\centering Given a description, generate SPCC.}
& \(\mathcal{A}_j,\mathcal{T}_j\) & \(\tilde{\mathcal{D}_j}\) \\
\midrule
\textit{Caption} & \parbox[c]{7.6cm}{\centering Given the CAD code, generate SPCC, which incorporates hierarchical descriptions.} &
\(C(\mathcal{D}_j)\) & \(\tilde{\mathcal{D}_j}\) \\
\midrule
\textit{Completion} & \parbox[c]{7.6cm}{\centering Given the partial SPCC, complete the missing part.} &
\(\text{Partial}(\tilde{\mathcal{D}_j})\) & \(\tilde{\mathcal{D}_j}\) \\
\midrule
\textit{Addition} & \parbox[c]{7.6cm}{\centering Given the CAD code and an instruction, add a specific component to the CAD model.} &
\(C({\mathcal{D}}_{j\setminus {\mathcal{P}_j^k}} )\) &
\(C(\mathcal{D}_j)\) \\
\midrule
\textit{Addition\textsuperscript{*}} & \parbox[c]{7.6cm}{\centering Given the SPCC and an instruction, add a specific component to the CAD model.} & 
${\tilde{\mathcal{D}}_{j\setminus {P_k}}}$
& \(\tilde{\mathcal{D}_j}\) \\
\midrule
\textit{Deletion} & \parbox[c]{7.6cm}{\centering Given the CAD code and an instruction to delete a specific component, remove the component.} &
\(C(\mathcal{D}_j)\) &
\(C({\mathcal{D}}_{j\setminus {\mathcal{P}_j^k}} )\) \\
\midrule
\textit{Deletion\textsuperscript{*}} & \parbox[c]{7.6cm}{\centering Given the SPCC and an instruction to delete a specific component, remove the component.} &
${\tilde{\mathcal{D}}_{j}}$
& \(\tilde{\mathcal{D}}_{j\setminus {P_k}} \) \\
\midrule
\end{tabular}
}
\caption{The overview of CAD-related tasks.}
\label{tab:graph_tasks}
\vspace{-6mm}
\end{table}
\vspace{-2.5mm}
\subsection{Training}\label{33-training}

\textbf{SPCC-Adaptive Pretraining} We select LLaMA3-8B \cite{dubey2024llama} as our foundational LLM and conduct
SPCC-adaptive pretraining on this LLM using the SPCC corpus. 
The traditional pretraining method creates input contexts by randomly concatenating pretraining data. However, in the same context, the preceding documents do not offer any assistance in predicting the content of the following document. Some CAD models have only minor differences, such as a change in the position of a component. To enable LLMs to capture these differences between similar CAD models for more efficient learning, similar to \cite{shi2023context}, we group similar CAD models together, so that each input context contains similar CAD models. Specifically, we use a pretrained CLIP \cite{radford2021learning} model to map each CAD model $\mathcal{D}_j\in \mathcal{D}$ to an embedding $\mathbf{E}({I}_j)$ based on its image ${I}_j$. Then, we calculate the similarity between pairs of CAD models using cosine similarity:
\vspace{-2.5mm}
\begin{equation}
s(\mathcal{D}_i,\mathcal{D}_j)=\text{cos}(\mathbf{E}(\mathcal{D}_i),\mathbf{E}(\mathcal{D}_j)).
\end{equation}
Subsequently, we construct a CAD document graph based on the similarities and build input contexts for pretraining by traversing this graph.
After deriving the final input contexts through the aforementioned methods, SPCC-adaptive pretraining optimizes a standard autoregressive language modeling objective, which maximizes the conditional probabilities of each token given its preceding tokens as context. Formally, given an input context represented by tokens $\mathcal{X}=\{x_0,x_1,\dots,x_{n-1},x_n\}$, CAD-Llama applies this objective by maximizing the following log-likelihood:
\vspace{-2.5mm}
\begin{equation}
\mathcal{L}(\mathcal{X})=\sum_{i=1}^{n}\text{log}P(x_i|x_{i-1},x_{i-2},\dots,x_0; \Phi),
\end{equation}
where $n$ is the context window size, $x_n$ is the special token \texttt{<|end\_of\_text|>}, and $\Phi$ denotes the model parameters.
After pretraining, the model is equipped with essential capabilities for generating and understanding SPCC, and we name this model CAD-Llama.

\textbf{CAD-centric Instruction Tuning} 
Given the CAD-centric multitask instructional dataset \(D = \{(X_i, Y_i)\}_{i=1}^{N}\),
where \(X_i\) represents the input along with the corresponding
instruction description, and \(Y_i\) represents the corresponding
output, we fine-tune CAD-Llama on this dataset, employing LoRA \cite{hu2021lora} for parameter-efficient tuning, with the objective
of maximizing the following log-likelihood:
\vspace{-2.5mm}
\begin{equation}
\mathcal{L}(D) = \sum_{i=1}^{N} \log P(Y_i \mid X_i;  \Theta ),\\
\end{equation}
where \(\Theta\) is the trainable parameters of CAD-Llama. After this process,
we obtained CAD-Llama-INS. The experiments in the following section
demonstrate that the CAD-related instruction tuning process enhances the
model\textquotesingle s performance on a series of downstream tasks.
\section{Experiments}\label{Experiments}

In this section, we present the details of the experiments and the experimental results to evaluate the performance of our proposed method.

\subsection{Experimental Setups}\label{experimental-setups}

\textbf{Datasets} In our experiment we adopt DeepCAD\cite{wu2021deepcad} dataset, which contains
approximately 178K parametric CAD models.
We observed that many simple CAD models (e.g., cubes) in the dataset may introduce repetitive patterns, potentially degrading performance \cite{kandpal2022deduplicating, lee2021deduplicating}. We removed most of this data and applied a similar de-duplication method from \cite{xu2023hierarchical, xu2022skexgen}, leaving approximately 100K CAD models for training.
The training data is processed using the method described in Section \ref{sec:spcc-sec} to obtain the
pretraining corpora for CAD-Llama. During the instruction tuning
phase, we sampled 12K entries from each task in the training set
to construct an instruction dataset, resulting in 48K
entries.

\textbf{Metrics} For unconditional generation, we used metrics from \cite{wu2021deepcad, xu2023hierarchical, xu2022skexgen, Ma2024DrawStep}, which include: (1) Coverage
(\textit{COV}) measures how well the
generative model covers the real data distribution. (2) Minimum Matching
Distance (\textit{MMD}) calculates the minimum distance between
generated samples and real samples. (3)
Jensen-Shannon Divergence (\textit{JSD}) quantifies the similarity between
the distributions. (4) The success ratio (\({S_R}\)) assesses the
proportion of valid generated CAD models. (5) The \textit{Novel} score quantifies the proportion of generated CAD sequences that do not appear in the training set.

For \textit{text-to-CAD} task, the metrics include: (1) the accuracy of CAD model
reconstructions \({ACC_{T}}\) \cite{Li2024CADTranslator}, consists of
command accuracy \({ACC_{cmd}}\), parameter
accuracy \({ACC_{param}}\) \cite{wu2021deepcad}, and success ratio
(\({S_R}\)), with these metrics combined to compute an overall accuracy:
\({ACC_{T}} = \frac{1}{2} \left( \frac{{ACC_{cmd}} + {ACC_{param}}}{2} + {S_R} \right)\)
(2) Median Chamfer Distance (\textit{MCD}). (3) \textit{MMD} and (4) \textit{JSD}.

For CAD captioning, we use \textit{BLEU} \cite{papineni2002bleu}, \textit{Rouge} \cite{lin2004rouge} to
measure the closeness of generated captions to reference captions. For
the deletion task, we use Exact Match (\textit{EM}) to evaluate whether the generated CAD model matches the ground truth. For the 
addition task, we use \({ACC_{cmd}}\) and
\({ACC_{param}}\) to evaluate the accuracy of the added modules.

\textbf{Implementation details} We select LLaMA3-8B-HF \cite{dubey2024llama} as our backbone.
The learning rate is set to 2e-5 with the AdamW optimizer \cite{loshchilov2017decoupled} , and
a linear learning rate warm-up is applied. The size of the context window is
2048 during SPCC-adaptive pretraining and 4096 during instruction
tuning. To improve training efficiency, we use DeepSpeed \cite{rasley2020deepspeed}, Flash-Attention \cite{dao2022flashattention}. Furthermore, we perform full fine-tuning during pretraining and use LoRA \cite{hu2021lora} for parameter-efficient training in instruction tuning, using a rank of 256 and a \textit{lora}\_\(\alpha\) value of 128.

\textbf{Baselines} We compare our method with a
series of baseline methods. For unconditional generation, this includes
parametric CAD generation models DeepCAD \cite{wu2021deepcad}, SkexGen \cite{xu2022skexgen} and HNC-CAD \cite{xu2023hierarchical}; 
For CAD-related downstream tasks, our baseline models include
the open-source LLMs LLaMA3-8B \cite{dubey2024llama} and Mistral-7B
\cite{jiang2023mistral}, as well as the proprietary models GPT-4 \cite{achiam2023gpt}
and GPT-3.5 \cite{ouyang2022training}.
For the text-to-CAD task, our baselines also include CAD-Translator \cite{Li2024CADTranslator} and Text2CAD \cite{khan2024text2cad}, both of which are based on the text-to-CAD transformer architecture.

\subsection{Unconditional Generation}\label{unconditional-generation}

\begin{figure}[t]
    \centering
	\includegraphics[width=\linewidth]{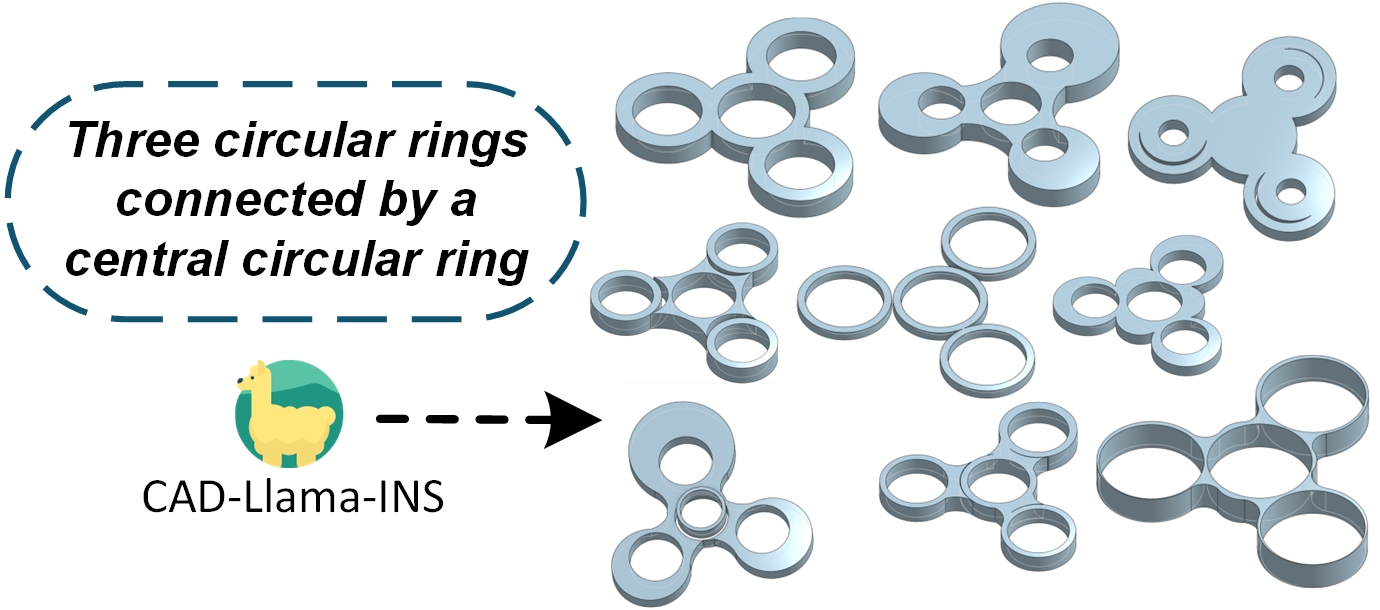}
    \vspace{-6mm}
	\caption{Qualitative example demonstrating CAD-Llama-INS generates CAD models that both conform to the description and exhibit diversity based on an abstract text description.}
 \label{fig:diversity}
\end{figure}

\begin{figure}[t]
	\includegraphics[width=\linewidth]{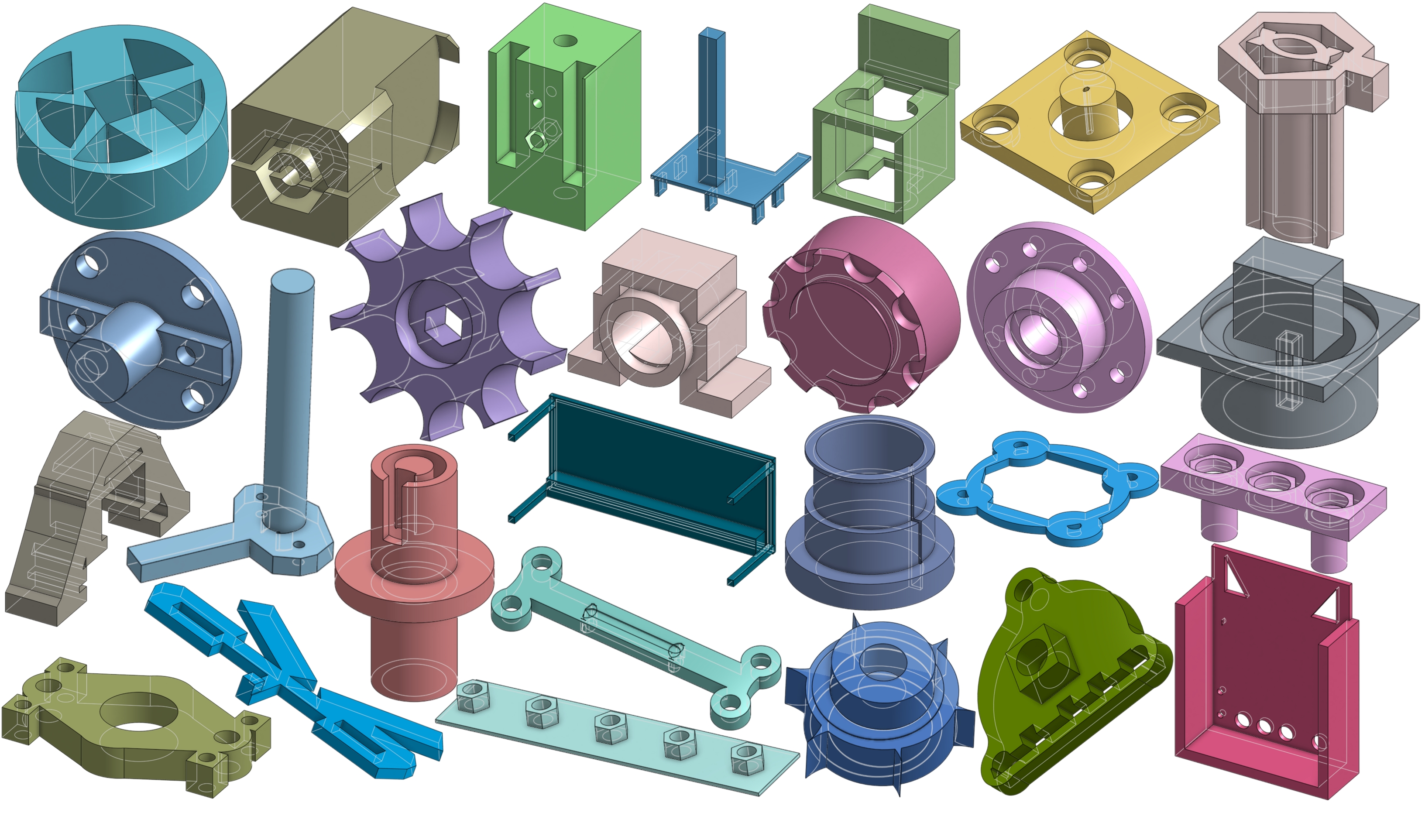}
    \vspace{-2mm}
	\caption{The unconditional generation results of CAD-Llama, demonstrate a wide range of complexity and diverse outputs.}
    \vspace{-3mm}
 \label{fig:uncondition_figure}
\end{figure}

We use the pretrained CAD-Llama for unconditional generation, prompted by the common prefix in SPCC format: \text{``}Description of the CAD model\text{''}. Each method generates 9,000 samples, with 2,000 points sampled for each one, which are compared to 3,000 randomly selected samples from the test set. Table \ref{tab:table_uncond} presents the main results on unconditional generation. For \textit{COV}, CAD-Llama achieves results comparable to HNC-CAD, indicating that after pretraining on the SCPP corpus, CAD-Llama has developed the capability to generate diverse CAD models. 
In \textit{MMD} and \textit{JSD}, CAD-Llama demonstrates superior performance with scores of 0.96 and 0.66, indicating a narrower distribution over the target space. For \({S_R}\), CAD-Llama achieves the highest value of 99.90, indicating highly stable results, surpassing the other three transformer-based methods, which exhibit significantly lower \({S_R}\) values. Figure \ref{fig:diversity} qualitatively illustrates that, given an abstract description, CAD-Llama-INS has the ability to generate CAD models that are both consistent with the text and diverse in nature, providing wide range of options and offering inspiration. Additionally, Figure \ref{fig:uncondition_figure} shows the unconditional generation results of CAD-Llama, demonstrating the model's ability to generate CAD models of varying complexity and diversity.

\begin{table}[t!]
\resizebox{\linewidth}{!}{
\begin{tabular}{lcccccc}
\toprule
\textbf{Method} & \textbf{COV} $\uparrow$ & \textbf{MMD} $\downarrow$ & \textbf{JSD} $\downarrow$ & $\mathbf{S_R}$ $\uparrow$ & \textbf{Novel} $\uparrow$ \\
\midrule
DeepCAD \cite{wu2021deepcad} & 66.68 & 1.19 & 2.59 & 61.84 & 91.7\\
SkexGen \cite{xu2022skexgen} & 77.42 & 1.07 & 0.93 & 72.26 & \textbf{99.1}\\
HNC-CAD \cite{xu2023hierarchical} & \textbf{80.46} & 0.98 & 0.74 & 79.11 & 93.9\\
\midrule
CAD-Llama (Ours) & 79.93 & \textbf{0.96} & \textbf{0.66} & \textbf{99.90} & 97.1\\
\bottomrule
\end{tabular}}
\caption{Results on unconditional generation. We present the test set \textit{Coverage} (\textbf{COV}) of generated CAD sequences, \textit{Minimum Matching Distance} (\textbf{MMD}), \textit{Jensen-Shannon Divergence} (\textbf{JSD}), \textit{Success Ratio} ($\mathbf{S_R}$) and \textbf{Novel} score. \textbf{COV}, $\mathbf{S_R}$ and \textbf{Novel} score are multiplied by 100\%. \textbf{JSD} and \textbf{MMD} are multiplied by $10^2$. $\uparrow$: the higher the better, $\downarrow$: the lower the better.}
\label{tab:table_uncond}
\vspace{-20pt}
\end{table}

\subsection{Main Results on Text-to-CAD Task}\label{main-results-on-text2cad-task}
\vspace{-2mm}
In the \textit{text-to-CAD} task, CAD-Llama-INS demonstrated superior performance compared to the transformer-based baseline methods, as well as GPT, LLaMA3, and others. As shown in Table \ref{tab:table_text2cad}, CAD-Llama-INS surpassed CAD-Translator and Text2CAD in accuracy by approximately 14\% and significantly outperformed other LLMs. This demonstrates the efficacy of our approach in leveraging 
LLMs to produce CAD models that more accurately reflect textual descriptions. Furthermore, our method demonstrated substantial improvements over the baselines in metrics such as \textit{MCD}, \textit{MMD}, and \textit{JSD}, indicating a closer geometric alignment with ground truth. These results underscore the limitations of transformer-based method, which, despite their ability to predict corresponding commands, often struggle with accurately predicting the parameters essential for the precision of parameterized CAD models.

\begin{table}[!htbp]
\resizebox{\linewidth}{!}{
\setlength{\tabcolsep}{1.pt}
\begin{tabular}{lcccccc}
\toprule
\textbf{Method} & \(\textbf{ACC}_{T}\) $\uparrow$ & \textbf{MCD} $\downarrow$ & \textbf{MMD} $\downarrow$ & \textbf{JSD} $\downarrow$ \\
\midrule
GPT-3.5 \cite{ouyang2022training} & 20.39 & 32.56 & 3.27 & 13.60 \\
GPT-4 \cite{achiam2023gpt} & 20.03 & 25.62 & 3.33 & 18.09 \\
LLaMA3 \cite{dubey2024llama} & 17.26 & 17.33 & 4.10 & 12.36 \\
Mistral \cite{jiang2023mistral} & 13.12 & 32.79 & 4.71 & 18.42 \\
Text2CAD \cite{khan2024text2cad} & 69.91 & 20.64 & 3.02 & 9.98 \\
CAD-Translator \cite{Li2024CADTranslator} & 70.36 & 21.29 & 2.94 & 10.92 \\
\midrule
CAD-Llama-INS (Ours) & \textbf{84.72} & \textbf{10.53} & \textbf{1.54} & \textbf{3.59} \\
\bottomrule
\end{tabular}
}
\caption{Results on \textit{text-to-CAD} task. The metric \(\textbf{ACC}_{T}\) is multiplied by 100\%. \textbf{MCD}, \textbf{MMD}, and \textbf{JSD} are multiplied by $10^2$.}
\label{tab:table_text2cad}
\vspace{-4mm}
\end{table}

\subsection{Main Results on CAD-related
 Downstream Tasks}\label{main-results-on-other-cad-related-tasks}
We evaluate CAD-Llama-INS and baselines on CAD-related tasks, with baselines in a two-shot setting.
The main results are presented in Table \ref{tab:all_instruct_results}.
CAD-Llama-INS achieved an average score of 63.58\%, surpassing GPT-4 by 15.7\% and outperforming LLaMA3 and Mistral by approximately 30\%. For all tested  CAD-related tasks, CAD-Llama-INS outperformed almost all baseline LLMs. This indicates that fine-tuning on SPCC corpus significantly enhances the understanding and generation capabilities of LLMs for CAD.

\begin{figure}[!h]
    \centering
	\includegraphics[width=\linewidth]{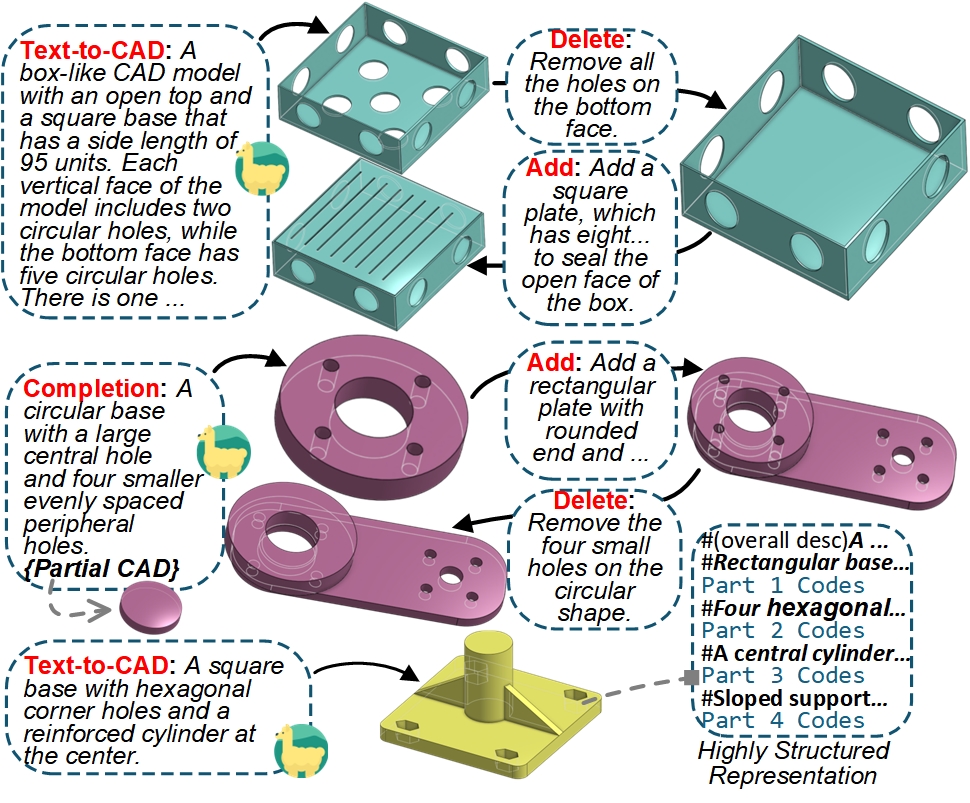}
	\caption{Examples of CAD-related tasks by using CAD-Llama-INS: the highly structured results with explicit annotations, with this series of tasks, aid designers in continuously optimizing the model, from initial construction to iterative refinement. For more detailed examples, please refer to the supplementary materials.}
 \label{fig:editing_flow_examples}
\end{figure}

For the \textit{deletion\textsuperscript{*}} and \textit{addition\textsuperscript{*}} tasks, CAD-Llama-INS significantly improved performance across all methods. Following structured annotation, GPT-4 leveraged its natural language reasoning capabilities to accurately identify modules for deletion, outperforming CAD-Llama-INS in the delete task. However, it struggles with the \textit{addition\textsuperscript{*}} task, which requires generating CAD parameters.
\begin{table}[t]
\centering
\setlength{\tabcolsep}{1pt}
\resizebox{\linewidth}{!}{
\begin{tabular}{c|c|cccc|c}
\toprule
\textbf{Tasks} & \textbf{Metric} & \textbf{GPT-4} & \textbf{GPT-3.5} & \textbf{LLaMA3} & \textbf{Mistral} &
\textbf{CAD-Llama-INS}\\
\hline
\multirow{3}{*}{\textit{Caption}}
& BLEU@1 & 27.34 & 25.03 & 22.41 & 21.26 & \textbf{43.44} \\
& BLEU@4 & 3.39 & 2.44 & 2.87 & 2.36 & \textbf{13.88} \\
& Rouge & 30.23 & 28.82 & 24.07 & 25.27 & \textbf{47.32} \\
\midrule
\multirow{2}{*}{\textit{Completion}}
& $\text{ACC}_{\text{cmd}}$ & 51.18 & 42.98 & 22.68 & 27.30 & \textbf{73.87} \\
& $\text{ACC}_{\text{param}}$ & 38.96 & 36.29 & 35.90 & 28.56 & \textbf{57.14} \\
\midrule
\multirow{2}{*}{\textit{Addition}}
& $\text{ACC}_{\text{cmd}}$ & 65.01 & 43.12 & 27.13 & 31.13 & \textbf{79.41} \\
& $\text{ACC}_{\text{param}}$ & 40.67 & 35.10 & 22.39 & 25.39 & \textbf{63.09} \\
\midrule
\multirow{2}{*}{\textit{Addition}\textsuperscript{*}}
& $\text{ACC}_{\text{cmd}}$ & $\underset{(+4.95)}{69.96}$ & $\underset{(+24.18)}{67.30}$ & $\underset{(+0.77)}{27.90}$ & $\underset{(+11.72)}{42.85}$ & $\underset{(+5.48)}{\textbf{84.89}}$ \\
& $\text{ACC}_{\text{param}}$ & $\underset{(+2.75)}{43.42}$ & $\underset{(+5.96)}{41.06}$ & $\underset{(+14.45)}{36.84}$ & $\underset{(+9.56)}{34.95}$ & $\underset{(+1.78)}{\textbf{64.87}}$ \\
\midrule
\textit{Deletion} & EM & 66.20 & 31.80 & 34.75 & 30.92 & \textbf{81.93} \\
\midrule
\textit{Deletion}\textsuperscript{*} & EM &  $\underset{(+24.21)}{\textbf{90.41}}$ & $\underset{(+21.23)}{53.03}$ & $\underset{(+8.94)}{43.69}$ & $\underset{(+8.89)}{39.81}$ & $\underset{(+7.62)}{89.55}$ \\
\midrule
Average &/&  47.88 & 36.99 & 27.33 & 28.16 & \textbf{63.58} \\
\bottomrule
\end{tabular}
}
\caption{Results (\%) on multiple CAD-related tasks. \textit{Deletion}\textsuperscript{*} and \textit{addition}\textsuperscript{*} indicate the results of first using CAD-Llama-INS to generate SPCC, followed by Delete and Add edits. More experimental results can be found in the supplementary materials.}
\label{tab:all_instruct_results}
\vspace{-10pt}
\end{table}
The experimental results indicate that SPCC offers a clear logical structure and semantic clarity, which improves the understanding and generation of LLMs. This also shows that CAD-Llama-INS, after pretraining on the SPCC corpus, effectively interprets intrinsic structural information. Figure \ref{fig:editing_flow_examples} presents two examples of \textit{text-to-CAD} and a range of CAD-related tasks using CAD-Llama-INS.
\begin{table}[t]
\resizebox{\linewidth}{!}{
\setlength{\tabcolsep}{1pt}
\begin{tabular}{lcccccc}
\toprule
\textbf{Method} & $\mathbf{ACC}_{\text{cmd}}\uparrow$ & $\mathbf{ACC}_{\text{param}}\uparrow$ & $\mathbf{S_R}\uparrow$ & \textbf{MCD}$\downarrow$ & \textbf{MMD}$\downarrow$ & \textbf{JSD}$\downarrow$\\
\midrule
\textit{SDCS} & 39.17 & 25.56 & 18.14 & 18.03 & 2.29 & 6.30 \\ 
\textit{SDCC} & 42.62 & 27.13 & 21.46 & 17.19 & 2.37 & 6.13 \\ 
\textit{SPCS} & 73.13 & 47.32 & 98.71 & 14.08 & 1.64 & 3.79 \\ 
\textit{SPCC} & \textbf{80.41} & \textbf{59.09} & \textbf{99.30} & \textbf{10.53} & \textbf{1.54} & \textbf{3.59} \\
\bottomrule
\end{tabular}
}
\caption{Evaluation of different CAD representation methods in the Text-to-CAD task. \textbf{SDCS} uses a single textual description as a prefix along with CAD command sequences, while \textbf{SDCC} uses CAD code with a single description. \textbf{SPCS} incorporates hierarchical descriptions with CAD command sequences, and \textbf{SPCC} combines hierarchical descriptions with CAD code.}
\vspace{-14pt}
\label{tab:table_ablation}
\end{table}
\subsection{Ablation Studies}\label{ablation-studies}
\vspace{-1mm}
Training data is crucial for the pretraining and fine-tuning of LLMs in this domain. We evaluate the impact of different representations of parametric CAD model training data on the \textit{text-to-CAD} task. The evaluation methods are categorized based on whether the CAD data is represented in code format or as its original command sequence, and whether hierarchical or single descriptions of the 3D shape and geometry are used:
(1) Single Description with CAD Sequences (\textit{SDCS}) uses CAD command sequences with a single-prefix description that encompasses both general details and components information.
(2) Single Description with CAD Code (\textit{SDCC}) uses CAD code
with the single-prefix description. (3) Structured Parametric
CAD Sequences (\textit{SPCS}) uses CAD command sequences with
hierarchical descriptions. (4) Structured Parametric CAD Code
(\textit{SPCC}) uses CAD code with hierarchical descriptions. For more details, please refer to the supplementary materials.

The experimental results in Table \ref{tab:table_ablation} show that the SPCC method
outperforms all other methods in the metrics, followed by the SPCS
method.
In contrast, the SDCS and SDCC methods, underperformed by approximately 30-40\% in \({ACC_{cmd}}\) and \({ACC_{param}}\). 
These findings highlight the significant advantage of using hierarchical descriptions in improving LLMs\textquotesingle{} ability to comprehend and generate CAD models, resulting in more accurate CAD generation. Additionally, representing CAD sequences in code format further enhances performance.
The structured CAD representation approach, which incorporates hierarchical descriptions, shows a significant high value \(S_R\), indicating a substantial increase in the stability of CAD
generation. In contrast, single-description methods are notably less
effective in generating valid CAD models.

\vspace{-2.5mm}
\section{Conclusion}
\label{Conclusion}
We introduce a novel paradigm that leverages the generative prior of LLMs into generating parametric CAD sequences. 
A hierarchical annotation pipeline is proposed to infuse textual descriptions of visual semantics and 3D shape at different levels of each CAD model via VLMs. 
A supervised fine-tuning paradigm is proposed to grant LLMs of general understanding and generation ability on parametric CAD models. 
An instruction tuning paradigm is proposed to fit into different downstream tasks of CAD model editing and operations. 
Experimental results show the superiority of our methods over traditional autoregressive methods as well as prevailing LLM baselines. 
In the future, with larger parameters and richer corpus, we believe that more exciting results of LLMs for CAD will appear.
\newline\newline
\noindent\textbf{Acknowledgment.} The computations in this research were performed using the CFFF platform of Fudan University.

{
    \small
    \bibliographystyle{ieeenat_fullname}
    \bibliography{main}

\begin{thebibliography}{67}
\providecommand{\natexlab}[1]{#1}
\providecommand{\url}[1]{\texttt{#1}}
\expandafter\ifx\csname urlstyle\endcsname\relax
  \providecommand{\doi}[1]{doi: #1}\else
  \providecommand{\doi}{doi: \begingroup \urlstyle{rm}\Url}\fi

\bibitem[Ope()]{OpenCASCADE}
Opencascade.
\newblock \url{https://www.opencascade.com/}.
\newblock Accessed: 20-Oct-2023.

\bibitem[Achiam et~al.(2023)Achiam, Adler, Agarwal, Ahmad, Akkaya, Aleman, Almeida, Altenschmidt, Altman, Anadkat, et~al.]{achiam2023gpt}
Josh Achiam, Steven Adler, Sandhini Agarwal, Lama Ahmad, Ilge Akkaya, Florencia~Leoni Aleman, Diogo Almeida, Janko Altenschmidt, Sam Altman, Shyamal Anadkat, et~al.
\newblock Gpt-4 technical report.
\newblock \emph{arXiv preprint arXiv:2303.08774}, 2023.

\bibitem[Badagabettu et~al.(2024)Badagabettu, Yarlagadda, and Farimani]{Badagabettu2024Query2CAD}
Akshay Badagabettu, Sai~Sravan Yarlagadda, and Amir~Barati Farimani.
\newblock {Query2CAD}: Generating {CAD} models using natural language queries.
\newblock \emph{arXiv preprint arXiv:2406.00144}, 2024.

\bibitem[Borrmann et~al.(2011)Borrmann, Elseberg, Lingemann, and N{\"u}chter]{borrmann20113d}
Dorit Borrmann, Jan Elseberg, Kai Lingemann, and Andreas N{\"u}chter.
\newblock The 3d hough transform for plane detection in point clouds: A review and a new accumulator design.
\newblock \emph{3D Research}, 2\penalty0 (2):\penalty0 1--13, 2011.

\bibitem[Brown(2020)]{brown2020language}
Tom~B Brown.
\newblock Language models are few-shot learners.
\newblock \emph{arXiv preprint arXiv:2005.14165}, 2020.

\bibitem[Bubeck et~al.(2023)Bubeck, Chandrasekaran, Eldan, Gehrke, Horvitz, Kamar, Lee, Lee, Li, Lundberg, Nori, Palangi, Ribeiro, and Zhang]{bubeck2023large}
Sébastien Bubeck, Venkat Chandrasekaran, Ronen Eldan, Johannes Gehrke, Eric Horvitz, Ece Kamar, Peter Lee, Yin~Tat Lee, Yuanzhi Li, Scott Lundberg, Harsha Nori, Hamid Palangi, Marco~Tulio Ribeiro, and Yi Zhang.
\newblock Large language models in medicine.
\newblock \emph{Nature Medicine}, 29:\penalty0 1936--1944, 2023.

\bibitem[Chabra et~al.(2020)Chabra, Lenssen, Ilg, Schmidt, Straub, Lovegrove, and Newcombe]{Chabra2020DeepLS}
Rohan Chabra, Jan~Eric Lenssen, Eddy Ilg, Tanner Schmidt, Julian Straub, Steven Lovegrove, and Richard Newcombe.
\newblock Deep local shapes: Learning local sdf priors for detailed 3d reconstruction.
\newblock In \emph{Proceedings of the European Conference on Computer Vision (ECCV)}, pages 608--625, 2020.

\bibitem[Collins et~al.(2023)Collins, Jiang, Frieder, Wong, Zilka, Bhatt, Lukasiewicz, Wu, Tenenbaum, Hart, Gowers, Li, Weller, and Jamnik]{collins2023evaluating}
Katherine~M. Collins, Albert~Q. Jiang, Simon Frieder, Lionel Wong, Miri Zilka, Umang Bhatt, Thomas Lukasiewicz, Yuhuai Wu, Joshua~B. Tenenbaum, William Hart, Timothy Gowers, Wenda Li, Adrian Weller, and Mateja Jamnik.
\newblock Evaluating language models for mathematics through interactions.
\newblock \emph{Proceedings of the National Academy of Sciences}, 120\penalty0 (24):\penalty0 e2318124121, 2023.

\bibitem[Dao et~al.(2022)Dao, Fu, Ermon, Rudra, and R{\'e}]{dao2022flashattention}
Tri Dao, Dan Fu, Stefano Ermon, Atri Rudra, and Christopher R{\'e}.
\newblock Flashattention: Fast and memory-efficient exact attention with io-awareness.
\newblock \emph{Advances in Neural Information Processing Systems}, 35:\penalty0 16344--16359, 2022.

\bibitem[Dubey et~al.(2024)Dubey, Jauhri, Pandey, Kadian, Al-Dahle, Letman, Mathur, Schelten, Yang, Fan, et~al.]{dubey2024llama}
Abhimanyu Dubey, Abhinav Jauhri, Abhinav Pandey, Abhishek Kadian, Ahmad Al-Dahle, Aiesha Letman, Akhil Mathur, Alan Schelten, Amy Yang, Angela Fan, et~al.
\newblock The llama 3 herd of models.
\newblock \emph{arXiv preprint arXiv:2407.21783}, 2024.

\bibitem[Duda and Hart(1972)]{duda1972use}
Richard~O Duda and Peter~E Hart.
\newblock Use of the hough transformation to detect lines and curves in pictures.
\newblock \emph{Communications of the ACM}, 15\penalty0 (1):\penalty0 11--15, 1972.

\bibitem[Ferruz and H{\"o}cker(2022)]{ferruz2022controllable}
Noelia Ferruz and Birte H{\"o}cker.
\newblock Controllable protein design with language models.
\newblock \emph{Nature Machine Intelligence}, 4\penalty0 (6):\penalty0 521--532, 2022.

\bibitem[Fischler and Bolles(1981)]{fischler1981random}
Martin~A Fischler and Robert~C Bolles.
\newblock Random sample consensus: a paradigm for model fitting with applications to image analysis and automated cartography.
\newblock \emph{Communications of the ACM}, 24\penalty0 (6):\penalty0 381--395, 1981.

\bibitem[Gao et~al.(2023)Gao, Wen, Gao, Wang, Zhang, and Lyu]{gao2023makes}
Shuzheng Gao, Xin-Cheng Wen, Cuiyun Gao, Wenxuan Wang, Hongyu Zhang, and Michael~R Lyu.
\newblock What makes good in-context demonstrations for code intelligence tasks with llms?
\newblock In \emph{2023 38th IEEE/ACM International Conference on Automated Software Engineering (ASE)}, pages 761--773. IEEE, 2023.

\bibitem[Gao(2023)]{gao2023learning}
Zhongpai Gao.
\newblock Learning continuous mesh representation with spherical implicit surface.
\newblock \emph{arXiv preprint arXiv:2301.04695}, 2023.

\bibitem[Hu et~al.(2021)Hu, Shen, Wallis, Allen-Zhu, Li, Wang, Wang, and Chen]{hu2021lora}
Edward~J Hu, Yelong Shen, Phillip Wallis, Zeyuan Allen-Zhu, Yuanzhi Li, Shean Wang, Lu Wang, and Weizhu Chen.
\newblock Lora: Low-rank adaptation of large language models.
\newblock \emph{arXiv preprint arXiv:2106.09685}, 2021.

\bibitem[Jiang et~al.(2023)Jiang, Sablayrolles, Mensch, Bamford, Chaplot, Casas, Bressand, Lengyel, Lample, Saulnier, et~al.]{jiang2023mistral}
Albert~Q Jiang, Alexandre Sablayrolles, Arthur Mensch, Chris Bamford, Devendra~Singh Chaplot, Diego de~las Casas, Florian Bressand, Gianna Lengyel, Guillaume Lample, Lucile Saulnier, et~al.
\newblock Mistral 7b.
\newblock \emph{arXiv preprint arXiv:2310.06825}, 2023.

\bibitem[Jiang et~al.(2024)Jiang, Xu, Dong, Chen, Ye, Yan, Ye, Zhang, Huang, and Zhang]{jiang2024hallucination}
Chaoya Jiang, Haiyang Xu, Mengfan Dong, Jiaxing Chen, Wei Ye, Ming Yan, Qinghao Ye, Ji Zhang, Fei Huang, and Shikun Zhang.
\newblock Hallucination augmented contrastive learning for multimodal large language model.
\newblock In \emph{Proceedings of the IEEE/CVF Conference on Computer Vision and Pattern Recognition}, pages 27036--27046, 2024.

\bibitem[Kandpal et~al.(2022)Kandpal, Wallace, and Raffel]{kandpal2022deduplicating}
Nikhil Kandpal, Eric Wallace, and Colin Raffel.
\newblock Deduplicating training data mitigates privacy risks in language models.
\newblock In \emph{International Conference on Machine Learning}, pages 10697--10707. PMLR, 2022.

\bibitem[Khan et~al.(2024)Khan, Sinha, Sheikh, Stricker, Ali, and Afzal]{khan2024text2cad}
Mohammad~Sadil Khan, Sankalp Sinha, Talha~Uddin Sheikh, Didier Stricker, Sk~Aziz Ali, and Muhammad~Zeshan Afzal.
\newblock Text2cad: Generating sequential cad models from beginner-to-expert level text prompts.
\newblock \emph{arXiv preprint arXiv:2409.17106}, 2024.

\bibitem[Koch et~al.(2019)Koch, Matveev, Jiang, Williams, Artemov, Burnaev, Alexa, Zorin, and Panozzo]{koch2019abc}
Sebastian Koch, Albert Matveev, Zhongshi Jiang, Francis Williams, Alexey Artemov, Evgeny Burnaev, Marc Alexa, Denis Zorin, and Daniele Panozzo.
\newblock Abc: A big cad model dataset for geometric deep learning.
\newblock In \emph{Proceedings of the IEEE/CVF conference on computer vision and pattern recognition}, pages 9601--9611, 2019.

\bibitem[Lambourne et~al.(2022)Lambourne, Willis, Jayaraman, Zhang, Sanghi, and Malekshan]{lambourne2022reconstructing}
Joseph~George Lambourne, Karl Willis, Pradeep~Kumar Jayaraman, Longfei Zhang, Aditya Sanghi, and Kamal~Rahimi Malekshan.
\newblock Reconstructing editable prismatic cad from rounded voxel models.
\newblock In \emph{SIGGRAPH Asia 2022 Conference Papers}, pages 1--9, 2022.

\bibitem[Lee et~al.(2021)Lee, Ippolito, Nystrom, Zhang, Eck, Callison-Burch, and Carlini]{lee2021deduplicating}
Katherine Lee, Daphne Ippolito, Andrew Nystrom, Chiyuan Zhang, Douglas Eck, Chris Callison-Burch, and Nicholas Carlini.
\newblock Deduplicating training data makes language models better.
\newblock \emph{arXiv preprint arXiv:2107.06499}, 2021.

\bibitem[Li et~al.(2023)Li, Guo, Zhang, and Yan]{li2023secad}
Pu Li, Jianwei Guo, Xiaopeng Zhang, and Dong-Ming Yan.
\newblock Secad-net: Self-supervised cad reconstruction by learning sketch-extrude operations.
\newblock In \emph{Proceedings of the IEEE/CVF Conference on Computer Vision and Pattern Recognition}, pages 16816--16826, 2023.

\bibitem[Li et~al.(2024{\natexlab{a}})Li, Song, Lou, and Zhou]{Li2024CADTranslator}
Xueyang Li, Yu Song, Yunzhong Lou, and Xiangdong Zhou.
\newblock {CAD Translator}: An effective drive for text to 3d parametric computer-aided design generative modeling.
\newblock In \emph{Proceedings of the 32nd ACM International Conference on Multimedia (MM 2024)}, Melbourne, Australia, 2024{\natexlab{a}}.

\bibitem[Li et~al.(2024{\natexlab{b}})Li, Sun, and Sha]{Li2024LLM4CAD}
Xingang Li, Yuewan Sun, and Zhenghui Sha.
\newblock {LLM4CAD}: Multi-modal large language models for 3d computer-aided design generation.
\newblock In \emph{Proceedings of the ASME 2024 International Design Engineering Technical Conferences \& Computers and Information in Engineering Conference (IDETC/CIE 2024)}, Washington, DC, USA, 2024{\natexlab{b}}.

\bibitem[Lin(2004)]{lin2004rouge}
Chin-Yew Lin.
\newblock Rouge: A package for automatic evaluation of summaries.
\newblock In \emph{Text summarization branches out}, pages 74--81, 2004.

\bibitem[Liu et~al.(2023)Liu, Lin, Li, Wang, Yacoob, and Wang]{liu2023aligning}
Fuxiao Liu, Kevin Lin, Linjie Li, Jianfeng Wang, Yaser Yacoob, and Lijuan Wang.
\newblock Aligning large multi-modal model with robust instruction tuning.
\newblock \emph{arXiv preprint arXiv:2306.14565}, 2023.

\bibitem[Liu et~al.(2024)Liu, Wang, Yang, Wang, Liu, Guo, and Xiao]{liu2024conversational}
Shengchao Liu, Jiongxiao Wang, Yijin Yang, Chengpeng Wang, Ling Liu, Hongyu Guo, and Chaowei Xiao.
\newblock Conversational drug editing using retrieval and domain feedback.
\newblock In \emph{Proceedings of the Twelfth International Conference on Learning Representations (ICLR)}, 2024.

\bibitem[Liu et~al.(2019)Liu, Tang, Lin, and Han]{Liu2019PVCNN}
Zhijian Liu, Haotian Tang, Yujun Lin, and Song Han.
\newblock Point-voxel {CNN} for efficient 3d deep learning.
\newblock In \emph{Advances in Neural Information Processing Systems (NeurIPS)}, pages 963--973, 2019.

\bibitem[Loshchilov(2017)]{loshchilov2017decoupled}
I Loshchilov.
\newblock Decoupled weight decay regularization.
\newblock \emph{arXiv preprint arXiv:1711.05101}, 2017.

\bibitem[Luo et~al.(2023)Luo, Sun, Xu, Zhao, Lou, Tao, Geng, Lin, Chen, and Zhang]{luo2023wizardmath}
Haipeng Luo, Qingfeng Sun, Can Xu, Pu Zhao, Jianguang Lou, Chongyang Tao, Xiubo Geng, Qingwei Lin, Shifeng Chen, and Dongmei Zhang.
\newblock Wizardmath: Empowering mathematical reasoning for large language models via reinforced evol-instruct.
\newblock \emph{arXiv preprint arXiv:2308.09583}, 2023.

\bibitem[Ma et~al.(2023{\natexlab{a}})Ma, Xu, Li, and Zhou]{Ma2023MultiCAD}
Weijian Ma, Minyang Xu, Xueyang Li, and Xiangdong Zhou.
\newblock {MultiCAD}: Contrastive representation learning for multi-modal 3d computer-aided design models.
\newblock In \emph{Proceedings of the 32nd ACM International Conference on Information and Knowledge Management (CIKM 2023)}, pages 1766--1776, 2023{\natexlab{a}}.

\bibitem[Ma et~al.(2024)Ma, Chen, Lou, Li, and Zhou]{Ma2024DrawStep}
Weijian Ma, Shuaiqi Chen, Yunzhong Lou, Xueyang Li, and Xiangdong Zhou.
\newblock Draw step by step: Reconstructing {CAD} construction sequences from point clouds via multimodal diffusion.
\newblock In \emph{Proceedings of the IEEE/CVF Conference on Computer Vision and Pattern Recognition (CVPR)}, pages 27154--27163, 2024.

\bibitem[Ma et~al.(2023{\natexlab{b}})Ma, Liu, Yu, Zhang, Jiang, Wang, and Li]{ma2023training}
Yingwei Ma, Yue Liu, Yue Yu, Yuanliang Zhang, Yu Jiang, Changjian Wang, and Shanshan Li.
\newblock At which training stage does code data help llms reasoning?
\newblock \emph{arXiv preprint arXiv:2309.16298}, 2023{\natexlab{b}}.

\bibitem[Madani et~al.(2023)Madani, Krause, Greene, Subramanian, Mohr, Holton, Olmos~Jr, Xiong, Sun, Socher, Fraser, and Naik]{madani2023large}
Ali Madani, Bryan Krause, Eric~R. Greene, Sandeep Subramanian, Benjamin~P. Mohr, James~M. Holton, Jose~L. Olmos~Jr, Ce Xiong, Zhongkai Sun, Richard Socher, James~S. Fraser, and Nikhil Naik.
\newblock Large language models generate functional protein sequences across diverse families.
\newblock \emph{Nature Biotechnology}, 41:\penalty0 25--33, 2023.

\bibitem[Maturana and Scherer(2015)]{Maturana2015VoxNet}
Daniel Maturana and Sebastian Scherer.
\newblock {VoxNet}: A 3d convolutional neural network for real-time object recognition.
\newblock In \emph{2015 IEEE/RSJ International Conference on Intelligent Robots and Systems (IROS)}, pages 922--928, 2015.

\bibitem[Ouyang et~al.(2022)Ouyang, Wu, Jiang, Almeida, Wainwright, Mishkin, Zhang, Agarwal, Slama, Ray, et~al.]{ouyang2022training}
Long Ouyang, Jeffrey Wu, Xu Jiang, Diogo Almeida, Carroll Wainwright, Pamela Mishkin, Chong Zhang, Sandhini Agarwal, Katarina Slama, Alex Ray, et~al.
\newblock Training language models to follow instructions with human feedback.
\newblock \emph{Advances in neural information processing systems}, 35:\penalty0 27730--27744, 2022.

\bibitem[Papineni et~al.(2002)Papineni, Roukos, Ward, and Zhu]{papineni2002bleu}
Kishore Papineni, Salim Roukos, Todd Ward, and Wei-Jing Zhu.
\newblock Bleu: a method for automatic evaluation of machine translation.
\newblock In \emph{Proceedings of the 40th annual meeting of the Association for Computational Linguistics}, pages 311--318, 2002.

\bibitem[Park et~al.(2019)Park, Florence, Straub, Newcombe, and Lovegrove]{Park_2019_CVPR}
Jeong~Joon Park, Peter Florence, Julian Straub, Richard Newcombe, and Steven Lovegrove.
\newblock {DeepSDF}: Learning continuous signed distance functions for shape representation.
\newblock In \emph{Proceedings of the IEEE/CVF Conference on Computer Vision and Pattern Recognition (CVPR)}, pages 165--174, 2019.

\bibitem[Qi et~al.(2017{\natexlab{a}})Qi, Su, Mo, and Guibas]{qi2017pointnet}
Charles~R Qi, Hao Su, Kaichun Mo, and Leonidas~J Guibas.
\newblock Pointnet: Deep learning on point sets for 3d classification and segmentation.
\newblock In \emph{Proceedings of the IEEE conference on computer vision and pattern recognition}, pages 652--660, 2017{\natexlab{a}}.

\bibitem[Qi et~al.(2017{\natexlab{b}})Qi, Yi, Su, and Guibas]{qi2017pointnet++}
Charles~Ruizhongtai Qi, Li Yi, Hao Su, and Leonidas~J Guibas.
\newblock Pointnet++: Deep hierarchical feature learning on point sets in a metric space.
\newblock \emph{Advances in neural information processing systems}, 30, 2017{\natexlab{b}}.

\bibitem[Qiu et~al.(2024)Qiu, Liu, Feng, Liu, Xiao, Collins, Tenenbaum, Weller, Black, and Sch{\"o}lkopf]{Qiu2024LLM_SGP}
Zeju Qiu, Weiyang Liu, Haiwen Feng, Zhen Liu, Tim~Z. Xiao, Katherine~M. Collins, Joshua~B. Tenenbaum, Adrian Weller, Michael~J. Black, and Bernhard Sch{\"o}lkopf.
\newblock Can large language models understand symbolic graphics programs?
\newblock \emph{arXiv preprint arXiv:2408.08313}, 2024.

\bibitem[Radford et~al.(2021)Radford, Kim, Hallacy, Ramesh, Goh, Agarwal, Sastry, Askell, Mishkin, Clark, et~al.]{radford2021learning}
Alec Radford, Jong~Wook Kim, Chris Hallacy, Aditya Ramesh, Gabriel Goh, Sandhini Agarwal, Girish Sastry, Amanda Askell, Pamela Mishkin, Jack Clark, et~al.
\newblock Learning transferable visual models from natural language supervision.
\newblock In \emph{International conference on machine learning}, pages 8748--8763. PMLR, 2021.

\bibitem[Rasley et~al.(2020)Rasley, Rajbhandari, Ruwase, and He]{rasley2020deepspeed}
Jeff Rasley, Samyam Rajbhandari, Olatunji Ruwase, and Yuxiong He.
\newblock Deepspeed: System optimizations enable training deep learning models with over 100 billion parameters.
\newblock In \emph{Proceedings of the 26th ACM SIGKDD International Conference on Knowledge Discovery \& Data Mining}, pages 3505--3506, 2020.

\bibitem[Ren et~al.(2022)Ren, Zheng, Cai, Li, and Zhang]{ren2022extrudenet}
Daxuan Ren, Jianmin Zheng, Jianfei Cai, Jiatong Li, and Junzhe Zhang.
\newblock Extrudenet: Unsupervised inverse sketch-and-extrude for shape parsing.
\newblock In \emph{European Conference on Computer Vision}, pages 482--498. Springer, 2022.

\bibitem[Riegler et~al.(2017)Riegler, Ulusoy, and Geiger]{Riegler2017OctNet}
Gernot Riegler, Ali~Osman Ulusoy, and Andreas Geiger.
\newblock {OctNet}: Learning deep 3d representations at high resolutions.
\newblock In \emph{Proceedings of the IEEE Conference on Computer Vision and Pattern Recognition (CVPR)}, pages 3577--3586, 2017.

\bibitem[Schnabel et~al.(2007)Schnabel, Wahl, and Klein]{schnabel2007efficient}
Ruwen Schnabel, Roland Wahl, and Reinhard Klein.
\newblock Efficient ransac for point-cloud shape detection.
\newblock In \emph{Computer graphics forum}, pages 214--226. Wiley Online Library, 2007.

\bibitem[Shen et~al.(2024)Shen, Li, Law, Atzmon, Fidler, Lucas, Gao, and Sharp]{spacemesh2024}
Tianchang Shen, Zhaoshuo Li, Marc Law, Matan Atzmon, Sanja Fidler, James Lucas, Jun Gao, and Nicholas Sharp.
\newblock Spacemesh: A continuous representation for learning manifold surface meshes.
\newblock In \emph{SIGGRAPH Asia 2024 Conference Papers (SA Conference Papers '24)}, page~11, New York, NY, USA, 2024. ACM.

\bibitem[Shi et~al.(2023)Shi, Min, Lomeli, Zhou, Li, Szilvasy, James, Lin, Smith, Zettlemoyer, et~al.]{shi2023context}
Weijia Shi, Sewon Min, Maria Lomeli, Chunting Zhou, Margaret Li, Gergely Szilvasy, Rich James, Xi~Victoria Lin, Noah~A Smith, Luke Zettlemoyer, et~al.
\newblock In-context pretraining: Language modeling beyond document boundaries.
\newblock \emph{arXiv preprint arXiv:2310.10638}, 2023.

\bibitem[Sun et~al.(2023)Sun, Shen, Cao, Liu, Li, Shen, Gan, Gui, Wang, Yang, et~al.]{sun2023aligning}
Zhiqing Sun, Sheng Shen, Shengcao Cao, Haotian Liu, Chunyuan Li, Yikang Shen, Chuang Gan, Liang-Yan Gui, Yu-Xiong Wang, Yiming Yang, et~al.
\newblock Aligning large multimodal models with factually augmented rlhf.
\newblock \emph{arXiv preprint arXiv:2309.14525}, 2023.

\bibitem[Uy et~al.(2022)Uy, Chang, Sung, Goel, Lambourne, Birdal, and Guibas]{uy2022point2cyl}
Mikaela~Angelina Uy, Yen-Yu Chang, Minhyuk Sung, Purvi Goel, Joseph~G Lambourne, Tolga Birdal, and Leonidas~J Guibas.
\newblock Point2cyl: Reverse engineering 3d objects from point clouds to extrusion cylinders.
\newblock In \emph{Proceedings of the IEEE/CVF Conference on Computer Vision and Pattern Recognition}, pages 11850--11860, 2022.

\bibitem[Wang et~al.(2024)Wang, Wu, Hou, Liu, Gao, and McAuley]{wang2024instructgraph}
Jianing Wang, Junda Wu, Yupeng Hou, Yao Liu, Ming Gao, and Julian McAuley.
\newblock Instructgraph: Boosting large language models via graph-centric instruction tuning and preference alignment.
\newblock \emph{arXiv preprint arXiv:2402.08785}, 2024.

\bibitem[Wang et~al.(2025)Wang, Chen, Le, Xu, Xu, Zhang, and Yang]{CAD-GPT}
Siyu Wang, Cailian Chen, Xinyi Le, Qimin Xu, Lei Xu, Yanzhou Zhang, and Jie Yang.
\newblock Cad-gpt: Synthesising cad construction sequence with spatial reasoning-enhanced multimodal llms.
\newblock In \emph{Proceedings of the AAAI Conference on Artificial Intelligence}, pages 7880--7888, 2025.

\bibitem[Willis et~al.(2021{\natexlab{a}})Willis, Jayaraman, Lambourne, Chu, and Pu]{willis2021engineering}
Karl~DD Willis, Pradeep~Kumar Jayaraman, Joseph~G Lambourne, Hang Chu, and Yewen Pu.
\newblock Engineering sketch generation for computer-aided design.
\newblock In \emph{Proceedings of the IEEE/CVF Conference on Computer Vision and Pattern Recognition}, pages 2105--2114, 2021{\natexlab{a}}.

\bibitem[Willis et~al.(2021{\natexlab{b}})Willis, Pu, Luo, Chu, Du, Lambourne, Solar-Lezama, and Matusik]{willis2021fusion}
Karl~DD Willis, Yewen Pu, Jieliang Luo, Hang Chu, Tao Du, Joseph~G Lambourne, Armando Solar-Lezama, and Wojciech Matusik.
\newblock Fusion 360 gallery: A dataset and environment for programmatic cad construction from human design sequences.
\newblock \emph{ACM Transactions on Graphics (TOG)}, 40\penalty0 (4):\penalty0 1--24, 2021{\natexlab{b}}.

\bibitem[Wong et~al.(2023)Wong, Guo, Hang, Ho, and Tan]{wong2023natural}
Man-Fai Wong, Shangxin Guo, Ching-Nam Hang, Siu-Wai Ho, and Chee-Wei Tan.
\newblock Natural language generation and understanding of big code for ai-assisted programming: A review.
\newblock \emph{Entropy}, 25\penalty0 (6):\penalty0 888, 2023.

\bibitem[Wu et~al.(2021)Wu, Xiao, and Zheng]{wu2021deepcad}
Rundi Wu, Chang Xiao, and Changxi Zheng.
\newblock Deepcad: A deep generative network for computer-aided design models.
\newblock In \emph{Proceedings of the IEEE/CVF International Conference on Computer Vision}, pages 6772--6782, 2021.

\bibitem[Wu et~al.(2023)Wu, Khasahmadi, Katz, Jayaraman, Pu, Willis, and Liu]{Wu2023CADLLM}
Sifan Wu, Amir Khasahmadi, Mor Katz, Pradeep~Kumar Jayaraman, Yewen Pu, Karl Willis, and Bang Liu.
\newblock {CAD-LLM}: Large language model for {CAD} generation.
\newblock In \emph{Proceedings of the Neural Information Processing Systems (NeurIPS) 2023}, 2023.

\bibitem[Xu et~al.(2024)Xu, Zhao, Wang, Liu, Ma, and Gao]{Cad-mllm}
Jingwei Xu, Zibo Zhao, Chenyu Wang, Wen Liu, Yi Ma, and Shenghua Gao.
\newblock Cad-mllm: Unifying multimodality-conditioned cad generation with mllm.
\newblock \emph{arXiv preprint arXiv:2411.04954}, 2024.

\bibitem[Xu et~al.(2021)Xu, Peng, Cheng, Willis, and Ritchie]{xu2021inferring}
Xianghao Xu, Wenzhe Peng, Chin-Yi Cheng, Karl~DD Willis, and Daniel Ritchie.
\newblock Inferring cad modeling sequences using zone graphs.
\newblock In \emph{Proceedings of the IEEE/CVF conference on computer vision and pattern recognition}, pages 6062--6070, 2021.

\bibitem[Xu et~al.(2022)Xu, Willis, Lambourne, Cheng, Jayaraman, and Furukawa]{xu2022skexgen}
Xiang Xu, Karl~DD Willis, Joseph~G Lambourne, Chin-Yi Cheng, Pradeep~Kumar Jayaraman, and Yasutaka Furukawa.
\newblock Skexgen: Autoregressive generation of cad construction sequences with disentangled codebooks.
\newblock In \emph{International Conference on Machine Learning}, pages 24698--24724. PMLR, 2022.

\bibitem[Xu et~al.(2023)Xu, Jayaraman, Lambourne, Willis, and Furukawa]{xu2023hierarchical}
Xiang Xu, Pradeep~Kumar Jayaraman, Joseph~G Lambourne, Karl~DD Willis, and Yasutaka Furukawa.
\newblock Hierarchical neural coding for controllable cad model generation.
\newblock \emph{arXiv preprint arXiv:2307.00149}, 2023.

\bibitem[Yu et~al.(2023)Yu, Jiang, Shi, Yu, Liu, Zhang, Kwok, Li, Weller, and Liu]{yu2023metamath}
Longhui Yu, Weisen Jiang, Han Shi, Jincheng Yu, Zhengying Liu, Yu Zhang, James~T Kwok, Zhenguo Li, Adrian Weller, and Weiyang Liu.
\newblock Metamath: Bootstrap your own mathematical questions for large language models.
\newblock \emph{arXiv preprint arXiv:2309.12284}, 2023.

\bibitem[Yuan et~al.(2024{\natexlab{a}})Yuan, Xu, Pan, Bousseau, Mitra, and Li]{yuan2024cadtalk}
Haocheng Yuan, Jing Xu, Hao Pan, Adrien Bousseau, Niloy~J Mitra, and Changjian Li.
\newblock Cadtalk: An algorithm and benchmark for semantic commenting of cad programs.
\newblock In \emph{Proceedings of the IEEE/CVF Conference on Computer Vision and Pattern Recognition}, pages 3753--3762, 2024{\natexlab{a}}.

\bibitem[Yuan et~al.(2024{\natexlab{b}})Yuan, Shi, and Huang]{Yuan2024OpenECAD}
Zhe Yuan, Jianqi Shi, and Yanhong Huang.
\newblock {OpenECAD}: An efficient visual language model for editable 3d-cad design.
\newblock \emph{Computers \& Graphics}, 124:\penalty0 104048, 2024{\natexlab{b}}.

\bibitem[Yue et~al.(2023)Yue, Qu, Zhang, Fu, Huang, Sun, Su, and Chen]{yue2023mammoth}
Xiang Yue, Xingwei Qu, Ge Zhang, Yao Fu, Wenhao Huang, Huan Sun, Yu Su, and Wenhu Chen.
\newblock {MAmmoTH}: Building math generalist models through hybrid instruction tuning.
\newblock \emph{arXiv preprint arXiv:2309.05653}, 2023.

\end{thebibliography}
}

\clearpage
\appendix
\renewcommand{\thesection}{\Alph{section}}
\setcounter{page}{1}
\maketitlesupplementary

\section{Overview}
\label{sec:overview}
In the supplementary material, we put forward some details about the data selection and method design. Cost analysis as well as extra experiment results are also put forward. The remaining parts are organized as follows.
\begin{itemize}
    \item First, we provide a cost analysis on both GPU resource and GPT-4o tokens. 
    \item Then we illustrate the format of our CAD code used throughout the pretraining and instruction tuning stage.
    \item After that we introduce the hierarchical annotation pipeline in detail with respect to CAD components, image extractor and two-stage prompting strategy.
    \item Finally we provide extra experiment results both quantitatively and qualitatively.
\end{itemize}
\section{Training Cost and GPT-4o Token Cost}
\label{sec:code_formatting}

Both SPCC-adaptive pretraining and instruction tuning stages are conducted on 4 A100 GPUs. Table \ref{tab:training_costs} summarizes the computational costs and token consumption for these stages. For generating finetuning data, during the SPCC-adaptive pretraining stage, altogether 70 million tokens are required to comprehend the image and generate prompts hierarchically. During the instruction tuning stage, 6 million tokens are used to generate instruction data. For the consumption of GPUs, SPCC-adaptive pretraining requires 38 A100-GPU hours and processes GPT-4o 70M tokens, while instruction tuning requires 12 A100-GPU hours and processes GPT-4o 6M tokens. This demonstrates that our model achieves efficient training with limited computational resources. Notably, during the Instruction Tuning phase, the model adapts effectively to various downstream tasks using only a small amount of data and training time.

\begin{table}[h]
\centering
\resizebox{\columnwidth}{!}{%
\begin{tabular}{c|c|c}
\toprule
\textbf{Stage} & \textbf{A100-GPU Hours} & \textbf{Tokens (GPT-4o)} \\ \midrule
SPCC-Adaptive Pretraining & 38 & 70M \\ \hline
Instruction Tuning & 12 & 6M \\ \bottomrule
\end{tabular}}
\caption{Training costs and token consumption during the two training stages. Tokens are used for prompt generation in each stage.}
\label{tab:training_costs}
\end{table}

\section{Details of CAD Code Formatting}
\label{sec:code_formatting}

We follow the annotations of DeepCAD \cite{wu2021deepcad} dataset and denote the components of the CAD command sequence. The complete set of command parameters is defined as \({p}_i = [x, y, \alpha, f, r, \theta, \phi, \gamma, p_x, p_y, p_z, s, e_1, e_2, b, u]\). We normalize and quantize these parameters as follows:
(1) For discrete coordinate parameters, including the sketch plane origin $(p_x, p_y, p_z)$, extrusion distances $(e_1, e_2)$, curve endpoint coordinates $(x, y)$, and the circle radius $r$, we quantize all continuous parameters into 256 levels, represent them with 8-bit integers, and recenter the origin from $(128, 128)$ to $(0, 0)$ for a more intuitive representation of scale.
(2) For angular parameters, including the sketch orientation angles $(\theta, \phi, \gamma)$ within the range $[-\pi, \pi]$ and the arc's sweep angle $\alpha$ within $[0, 2\pi]$, we use discrete values within the ranges $[-180, 180]$ and $[0, 360]$ degrees, respectively.
(3) The sketch profile scale $s$ is constrained within the range $[0, 2]$, while the boolean operation type $b$ can take one of the following values: \emph{new body}, \emph{join}, \emph{cut}, or \emph{intersect}. The extrusion type $u$ denotes one of three configurations: \emph{one-sided}, \emph{symmetric}, or \emph{two-sided}. These parameters are utilized in their original forms.
(4) The arc's counterclockwise flag $f$ is a binary indicator, which we represent as either True or False.

For converting the annotation of CAD construction sequence into a LLM-friendly format, we further extract the hierarchy of CAD construction sequences and organize them into python-like pseudocode. In particular, the SOL and EOS commands are abstracted as an object \texttt{Loop()} and an ending comment \texttt{\# End of code}, respectively. Other commands, such as \textit{Arc}, \textit{Line}, \textit{Circle}, and \textit{Extrude}, are represented as function calls with corresponding parameters as inputs of the function. Detailed examples are illustrated in Figure \ref{fig:SPCC_0} and \ref{fig:SPCC_1}.

\section{Details of Hierarchical Annotation Pipeline}
\label{sec:annotation_details}

\begin{figure*}[!h]
    \centering
	\includegraphics[width=\linewidth]{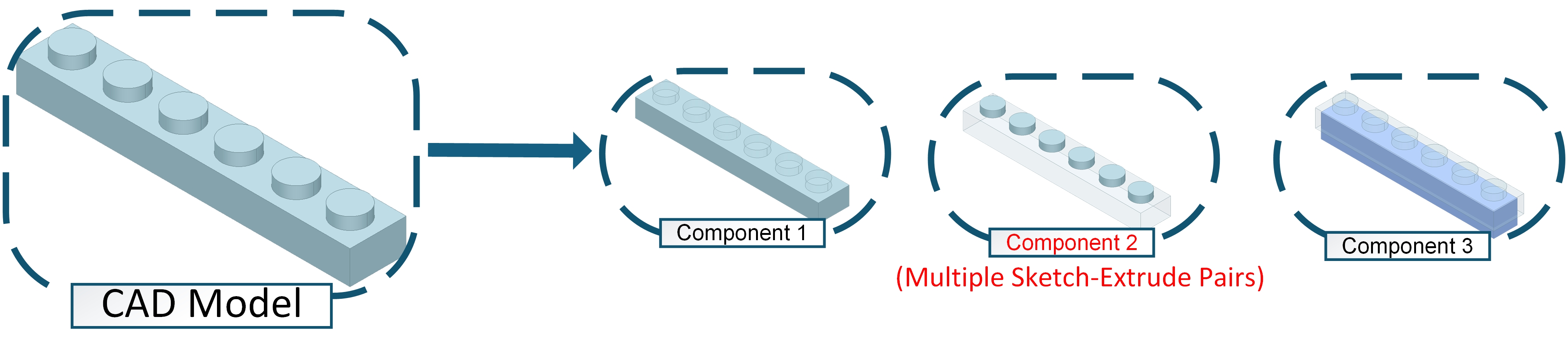}
	\caption{Illustration of defining a single component from consecutive equivalent sketch-extrude pairs based on specified criteria. Note that a large proportion of bottom cube in the final CAD model has been cut out, as is shown in component 3.}
 \label{fig:define_component}
\end{figure*}

\subsection{Definition of CAD Component}
\label{subsec:define_component}
In our definition, a CAD model consists of one or more components. Typically, a single sketch-extrude pair is treated as an individual component. However, when multiple identical sketch-extrude pairs occur consecutively in a CAD sequence, such as 10 cylinders uniformly distributed in a circular arrangement, describing each pair individually leads to redundancy and poses challenges for vision-language models (VLMs) in accurately capturing such repetitive structures.

To address this, when identical sketch-extrude pairs occur consecutively and their count exceeds a specified threshold (set to 3 in our experiments), we collectively define them as a single component. Otherwise, each sketch-extrude pair is treated as an individual component.
The equivalence of two sketch-extrude pairs is determined based on the following criteria: all commands and parameters must match, except for the sketch plane origin parameters $(p_x, p_y, p_z)$. As illustrated in Figure \ref{fig:define_component}, the second component comprises multiple sketch-extrude pairs.

\subsection{Annotation Image Generation Pipeline}
\label{Image_Generation_Pipeline}
The hierarchical annotation pipeline contains two stages. Different images are fed into the VLM in different stages. 
We propose two kinds of image extractors which extracts different features of the CAD model, namely \textbf{Components Images Extractor} and \textbf{Outlines Images Extractor}, as shown in Figure 3 in the main paper. Both of them are python scripts rendering CAD construction sequences using PythonOCC \cite{OpenCASCADE} (Python version of OCCT) while focusing on different aspects of a single model.
Taking the \textit{i}-th component of the \textit{j}-th CAD model $\mathcal{D}_j$ as an example, in the first stage, we use the \textbf{Components Images Extractor} and obtain the component image ${I}_j^i$ and its corresponding 2D sketch image ${\hat{I}}_j^i$. Specifically, ${I}_j^i$ is rendered from the default viewpoint by extracting the component's CAD command, while ${\hat{I}}_j^i$ is obtained by rendering the corresponding 2D sketch commands.
In the second stage, we use the \textbf{Outlines Images Extractor} and obtain the outline image ${\dot{I}}_j^i$, achieved by increasing the transparency of other components (set to 0.85 in our experiments) while keeping the target component's transparency unchanged during rendering; if the target component is used for \textit{cutting}, it is rendered in blue, as illustrated by component 3 in Figure \ref{fig:define_component}.

\subsection{Two Stage Prompting Methods.}
In this part we provide the detailed prompt used in Section \ref{Image_Generation_Pipeline}. In particular, two prompts are adopted where \textbf{prompt1} is for obtaining descriptions of individual components and \textbf{prompt2} is for acquiring both overall descriptions and component names. In \textbf{prompt1}, to enable GPT-4o to generate more detailed descriptions, we provide additional information that includes extrusion direction, extrusion length, and quantity information. The extrusion direction is included only when the CAD model is extruded in a specific direction, such as up, down, left, right, front, or back. We observed that over 95\% of the extrusion directions in the DeepCAD \cite{wu2021deepcad} dataset fall within these categories. Quantity information is added only when a component contains multiple sketch-extrude pairs (see Section \ref{subsec:define_component}), which helps mitigate the hallucination phenomenon in VLM. The specific content of the two prompts is as follows:

\textbf{Prompt1}: \textit{Background: The user now has a CAD model, which is formed by extruding a sketch. User input: The user will input two pictures, the first is the sketch, and the second is the CAD model after the sketch is extruded. Task: Describe the CAD model. Please describe the sketch in detail first, include the additional information in the description and output the final description result as a single line. Additional information: \{Extrusion direction and length information, Number information\} Examples: \{Two Description Example\}}

\textbf{Prompt2}: \textit{A CAD model may consist of multiple modules. Each module constitutes a part of the model, which can be a solid or a feature used for cutting, such as creating a hole. The user has a CAD consisting of \{num\_parts\} modules. The user will input \{num\_parts+1\} pictures, the first image is the original CAD model, followed by \{num\_parts\} images where each module is rendered with enhanced highlighting. These modules collectively form the original CAD model. Modules used for cutting are highlighted in blue. The subsequent description explains each of the four modules individually, following the order presented in the module images: \{Component Descriptions\} Task: You need to output three lines, Line 1: A concise description of the overall macro of CAD based on first image. Line 2: A detailed description that includes the specific characteristics of each of the \{num\_parts\} modules mentioned above, as well as the process by which they are assembled based on all provided images and component descriptions. Line 3: Short names for \{num\_parts\} modules. Example: \{Two Description Example\}}
\section{Details of Experiments}
\label{sec:exp_details}

\subsection{Prompts Used in Baseline Methods}
We provide the detailed prompt used in the baseline methods (GPT-4, GPT-3.5, LLaMA3, and Mistral) in Figure \ref{fig:GPT_prompts}, where \texttt{\{task\_definition\}} specifies the task instructions.

\subsection{Ablation Study Details in Main Experiment}
In the ablation study of the main experiment, we explored the impact of using different CAD representations for pretraining on the Text-to-CAD task. The evaluation methods are categorized based on whether the CAD data is in code format or raw sequences, and whether hierarchical or single descriptions are used. The single description \(\mathcal{SD}\) of the CAD model \(\mathcal{D}_j\) is defined as:
\[
\mathcal{SD}=\text{concat}\{\mathcal{A}_j,\mathcal{T}_j,\text{“Parts description:"},D_j\}
\]
where $D_j=\text{concat}\{\{\mathcal{S}_{j}^1,\mathcal{I}_{j}^1\},\{\mathcal{S}_{j}^2,\mathcal{I}_{j}^2\},\dots,\{\mathcal{S}_{j}^k,\mathcal{I}_{j}^k\}\}$
represent the full description of all \emph{k} components of \(\mathcal{D}_j\).

\begin{table}[h]
\centering
\resizebox{\linewidth}{!}{
\begin{tabular}{c|c|c|c }
\toprule
\textbf{Tasks} & \textbf{Metric} & \textbf{\textit{w/o} ICP} &
\textbf{\textit{with} ICP}\\
\hline
\textbf{Text-to-CAD}
& \(\text{ACC\(_{\text{cmd}}\)}\) & 79.89 & \textbf{80.41} \\
& \(\text{ACC\(_{\text{param}}\)}\) & 59.04 & \textbf{59.09} \\
\midrule
\textbf{Add}
& \(\text{ACC\(_{\text{cmd}}\)}\) & 77.73 & \textbf{79.41} \\
& \(\text{ACC\(_{\text{param}}\)}\) & 62.16 & \textbf{63.09} \\
\midrule
\textbf{Delete} & EM & 80.91 & \textbf{81.93} \\
\bottomrule
\end{tabular}
}
\caption{Performance comparison of CAD-related tasks with and without In-Context Pretraining (ICP). The results show that ICP improves performance across all tasks}
\label{tab:icl}
\end{table}
\subsection{Ablation Studies on Pretraining Method}
This section presents a simple ablation study on several CAD-related tasks to validate the effectiveness of In-Context Pretraining (ICP) \cite{shi2023context} in enhancing CAD-Llama-INS performance on downstream tasks. ICP is a method that groups related documents within the same input context, encouraging LLMs to read and reason across document boundaries. Similar to \cite{shi2023context}, we used a pretrained CLIP \cite{radford2021learning} model to encode CAD images and group similar CADs for pretraining based on their cosine similarity.

As shown in Table \ref{tab:icl}, ICP enhances the performance of downstream editing tasks, such as add and delete, by enabling LLMs to better capture the distinctions between different CAD structures during pretraining. This improved understanding allows the model to more effectively handle precise modifications required in these tasks. Additionally, ICL contributes to a marginal improvement in the Text-to-CAD task. 

\begin{table}[h]
\centering
\resizebox{\columnwidth}{!}{%
\begin{tabular}{c|cccc}
\toprule
{\textbf{Dataset}} & \(\textbf{ACC}_{T}\)$\uparrow$ & \textbf{MCD}$\downarrow$ & \textbf{MMD}$\downarrow$ & \textbf{JSD}$\downarrow$ \\ 
\midrule
DeepCAD& 84.72& 10.53& 1.54& 3.59\\ 
\hline
Fusion360 & 78.35& 23.06& 1.98& 6.01\\ 
\bottomrule
\end{tabular}}
\caption{Our model, CAD-Llama-INS, trained exclusively on the DeepCAD dataset, demonstrates strong generalization capabilities on the Fusion360 dataset in the Text-to-CAD task.}
\label{tab:cross_dataset}
\end{table}
\subsection{Cross Dataset Generalization}
To further evaluate the generalization ability of CAD-Llama-INS, we conducted experiments on the test set of the Fusion 360 \cite{willis2021fusion} dataset for the Text-to-CAD task. Similar to DeepCAD \cite{wu2021deepcad}, the Fusion 360 dataset also contains CAD construction sequences. We employed the hierarchical annotation pipeline to generate descriptions for the Fusion 360 dataset. These descriptions are used to prompt CAD-Llama-INS, which was pre-trained and fine-tuned exclusively on the DeepCAD dataset, to produce corresponding CAD models. The experimental results, as shown in Table \ref{tab:cross_dataset}, demonstrate that CAD-Llama-INS achieves strong generalization performance, achieving comparable or superior results on the Fusion 360 dataset despite being trained solely on DeepCAD. This highlights the effectiveness of our approach in adapting to new datasets. A qualitative analysis is also conducted, as illustrated in Figure \ref{fig:360_result}. Based on textual prompts, CAD-Llama-INS demonstrates the capability to generate CAD models that closely align with the ground truth.

\subsection{Qualitative results}
To comprehensively evaluate the performance of our approach, we provide qualitative results across multiple tasks. Specifically, qualitative results for text-to-CAD generation are illustrated in Figure \ref{fig:text2cad_0} to Figure \ref{fig:working_example_4}. Results for captioning tasks are presented in Figure \ref{fig:caption}, while results for unconditional generation are shown in Figure \ref{fig:uncond_res}. Additionally, results for multi-task evaluation, encompassing the process from initial construction to iterative refinement, are shown in Figures \ref{fig:working_example_0} to \ref{fig:working_example_4}.

\subsection{Examples of Failure Cases}
Our experimental results also show some limitations of our method, in some cases there are parameter generation errors and mismatching between the input text instruction and the generated CAD command sequences,  Figure \ref{fig:fail_case} illustrates some failure cases of Text-to-CAD generation.

\begin{figure*}[h]
    \centering
	\includegraphics[width=\linewidth]{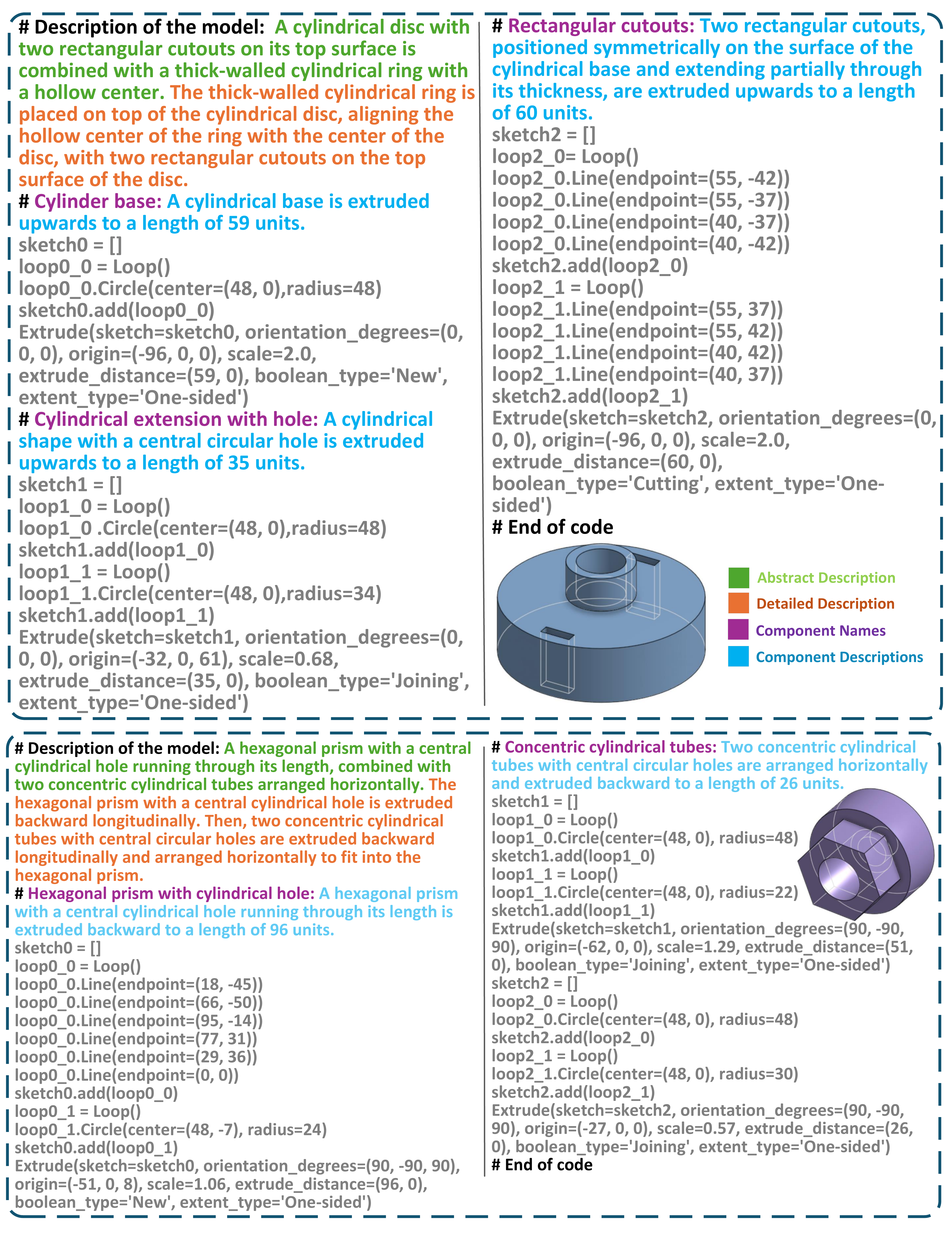}
	\caption{Examples of SPCC data representation, generated by our CAD-Llama-INS.}
 \label{fig:SPCC_0}
\end{figure*}

\begin{figure*}[h]
    \centering
	\includegraphics[width=\linewidth]{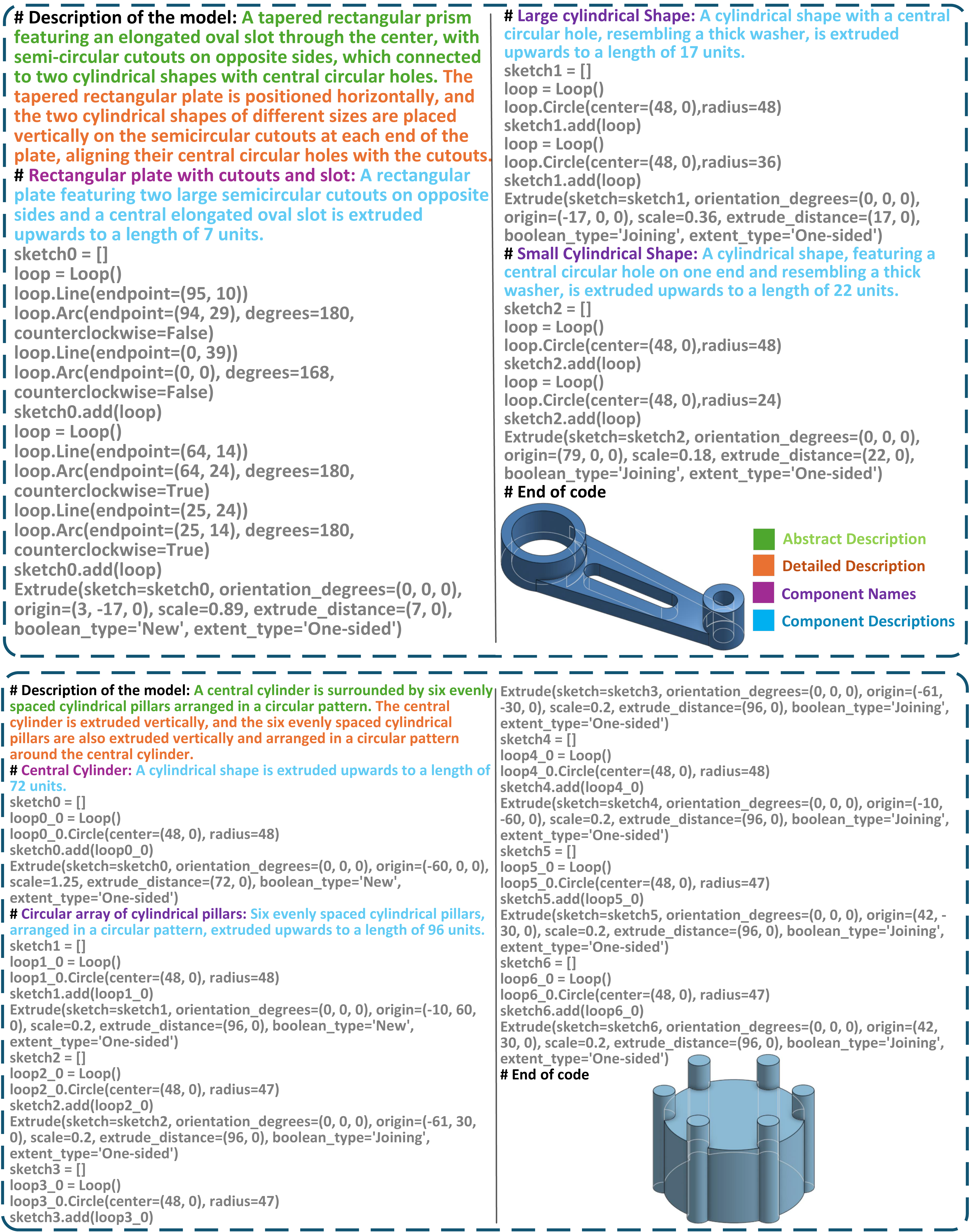}
	\caption{Examples of SPCC data representation, generated by our CAD-Llama-INS.}
 \label{fig:SPCC_1}
\end{figure*}

\begin{figure*}[h]
    \centering
	\includegraphics[width=\linewidth]{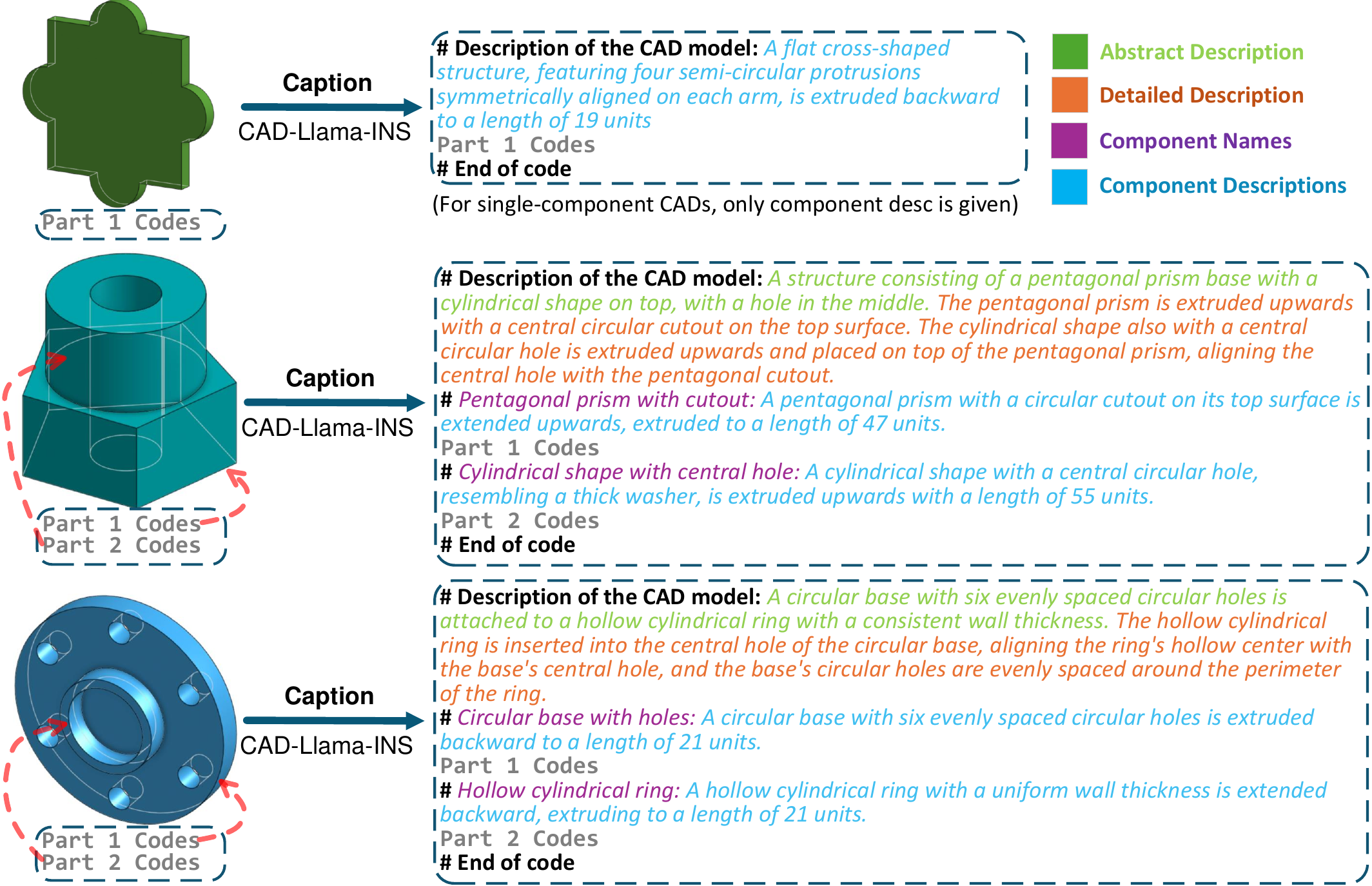}
	\caption{Examples of results from the Caption task, demonstrating the capabilities of CAD-Llama-INS in understanding the internal structure of raw CAD code and its geometric shapes.}
 \label{fig:caption}
\end{figure*}

\begin{figure*}[h]
    \centering
	\includegraphics[width=\linewidth]{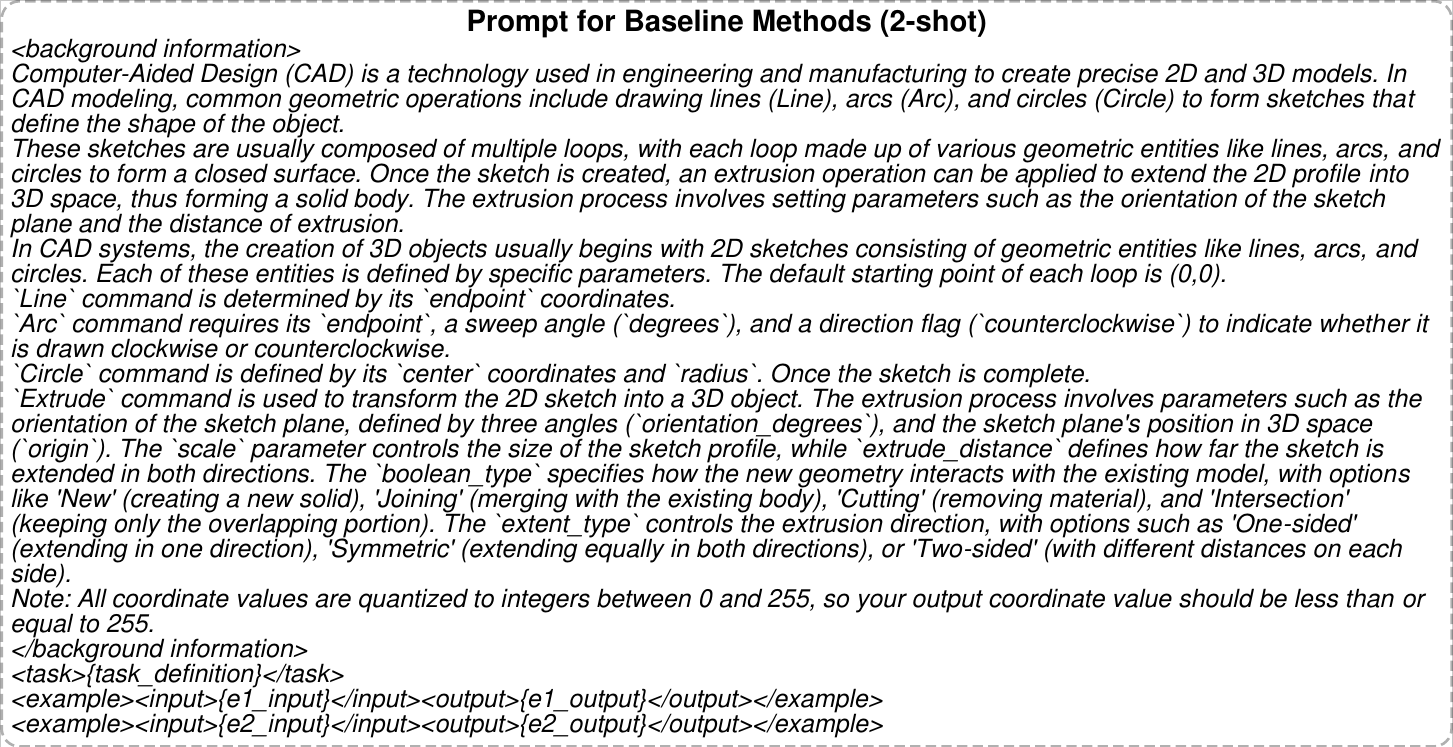}
	\caption{Detailed prompt used in the baseline methods (GPT-4, GPT-3.5, LLaMA and Mistrial).}
 \label{fig:GPT_prompts}
\end{figure*}

\begin{figure*}[h]
    \centering
    \vspace{-10mm}
	\includegraphics[width=\linewidth]{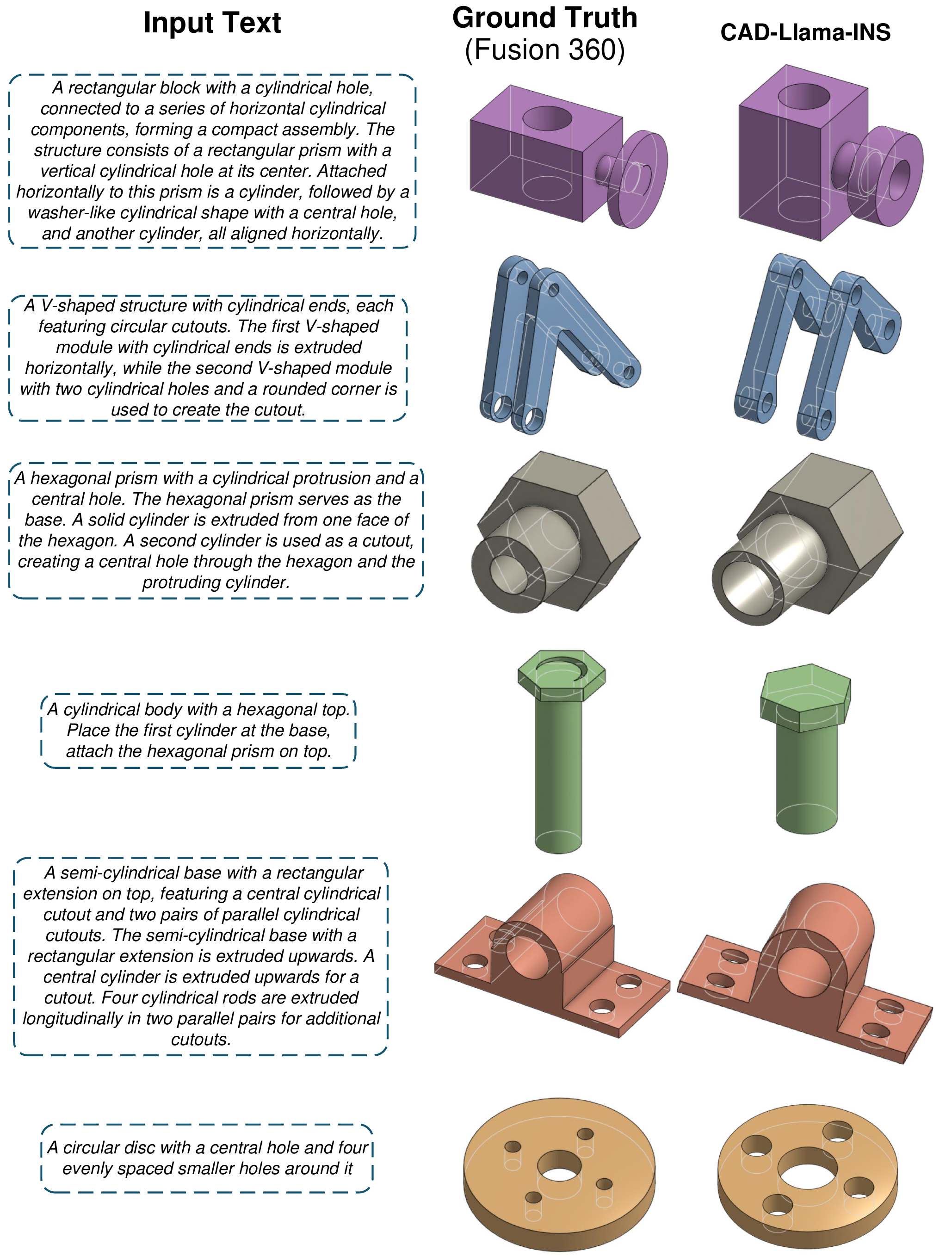}
	\caption{Comparison results of Text-to-CAD task on the Fusion 360 dataset.}
 \label{fig:360_result}
\end{figure*}

\begin{figure*}[h]
    \centering
    \vspace{-10mm}
	\includegraphics[width=\linewidth]{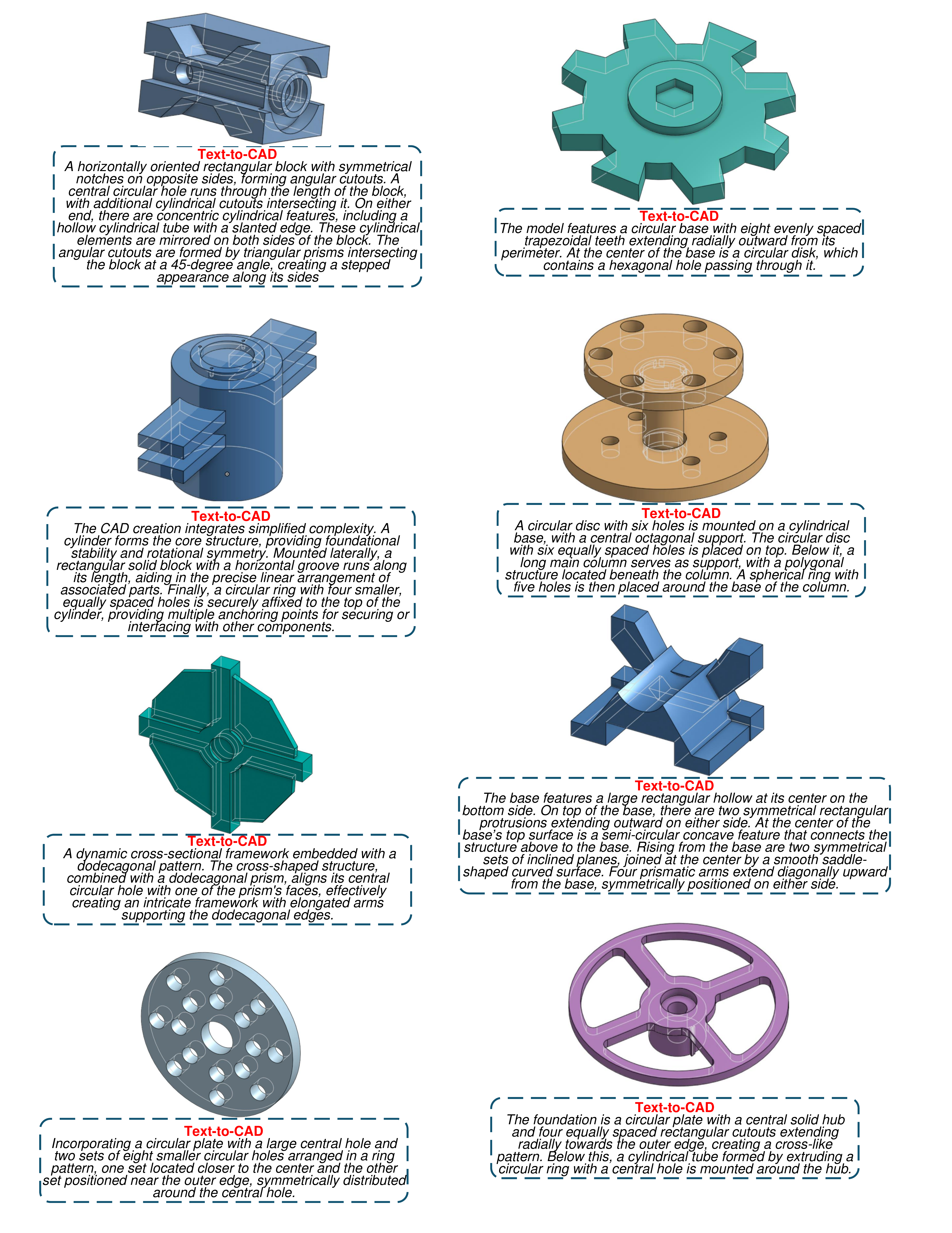}
	\caption{Supplementary results of the Text-to-CAD task generated by CAD-Llama-INS based on text prompts.}
 \label{fig:text2cad_0}
\end{figure*}

\begin{figure*}[h]
    \centering
    \vspace{-10mm}
	\includegraphics[width=\linewidth]{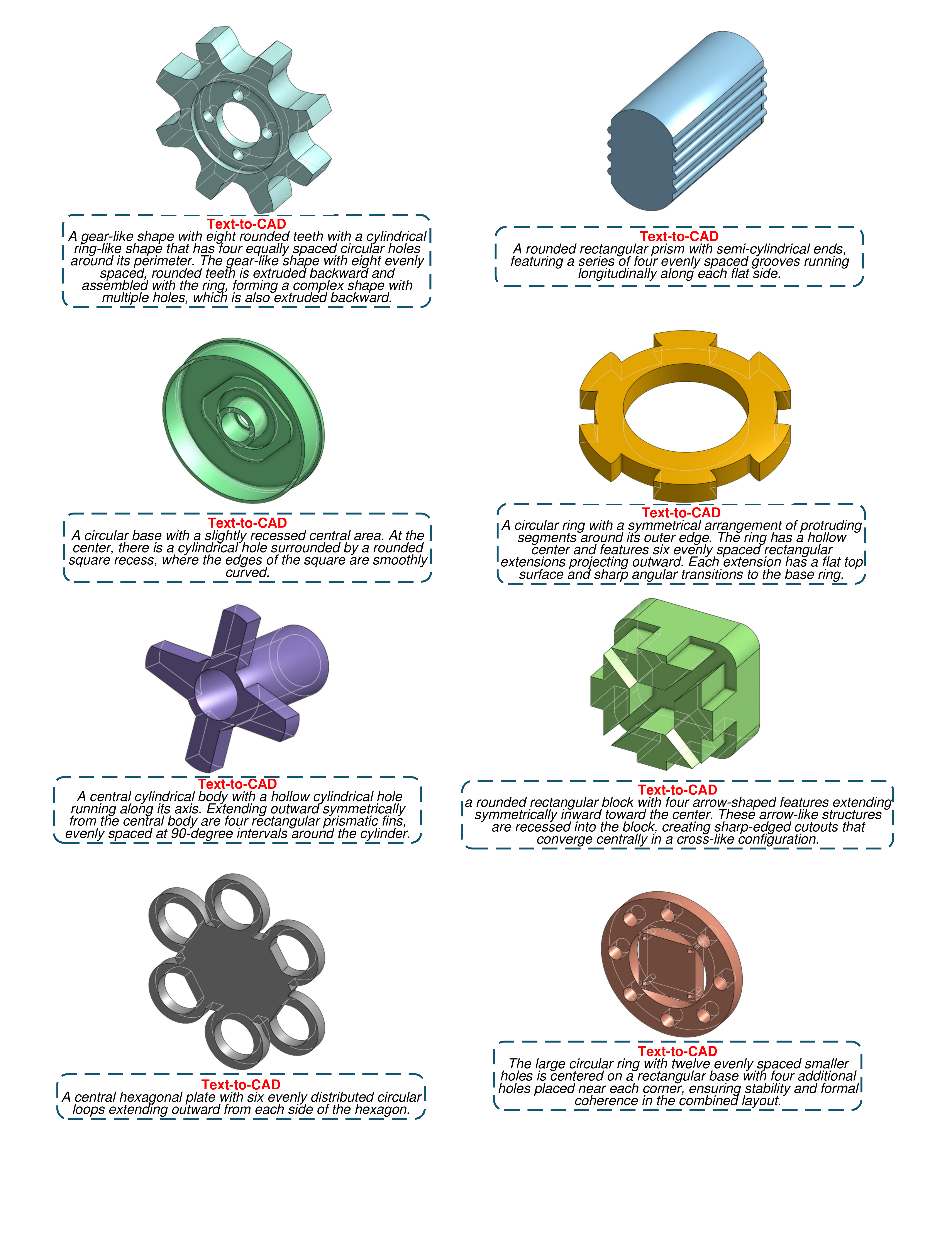}
	\caption{Supplementary results of the Text-to-CAD task generated by CAD-Llama-INS based on text prompts.}
 \label{fig:text2cad_1}
\end{figure*}

\begin{figure*}[h]
    \centering
    \vspace{-10mm}
	\includegraphics[width=\linewidth]{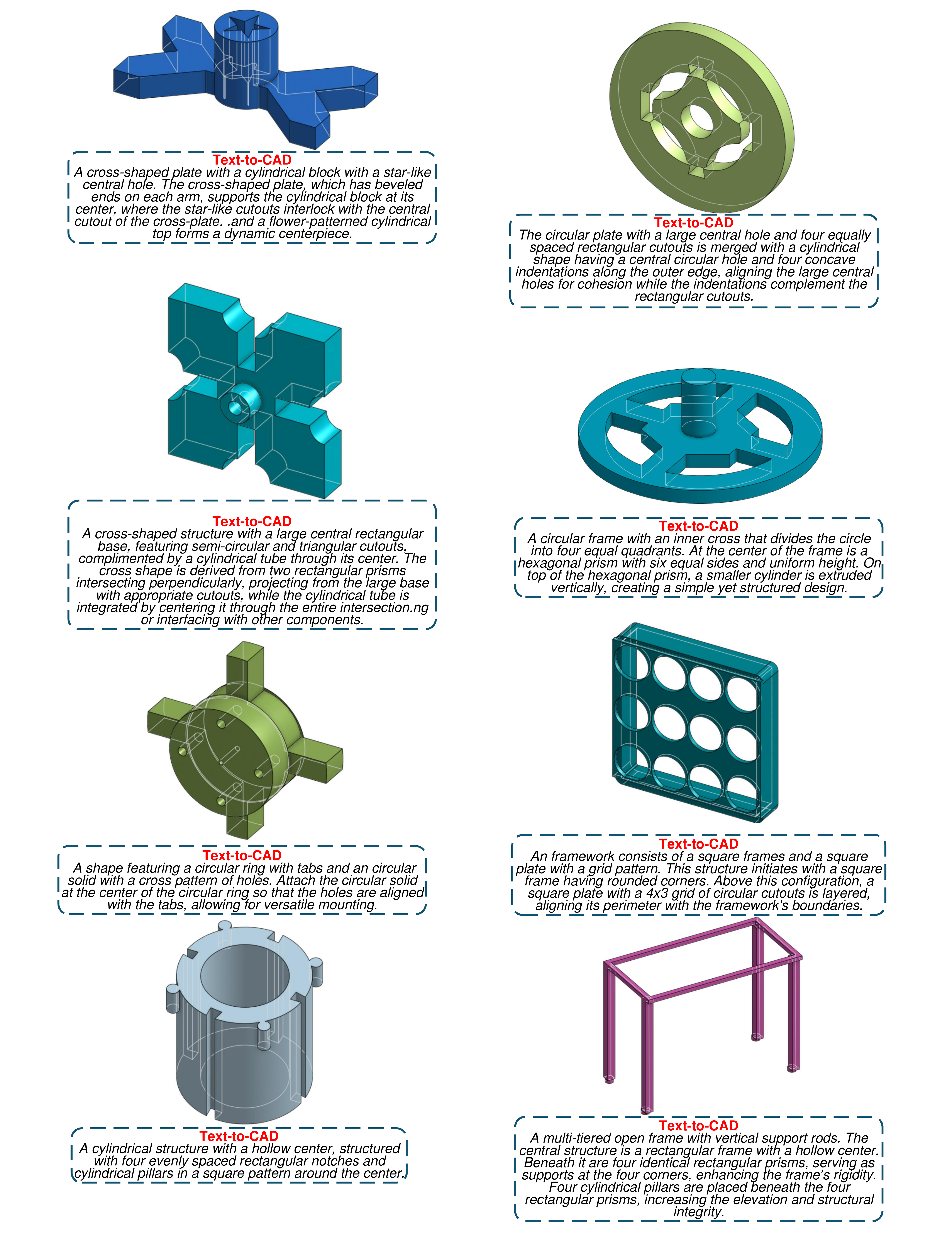}
	\caption{Supplementary results of the Text-to-CAD task generated by CAD-Llama-INS based on text prompts.}
 \label{fig:text2cad_2}
\end{figure*}

\begin{figure*}[h]
    \centering
    \vspace{-10mm}
	\includegraphics[width=\linewidth]{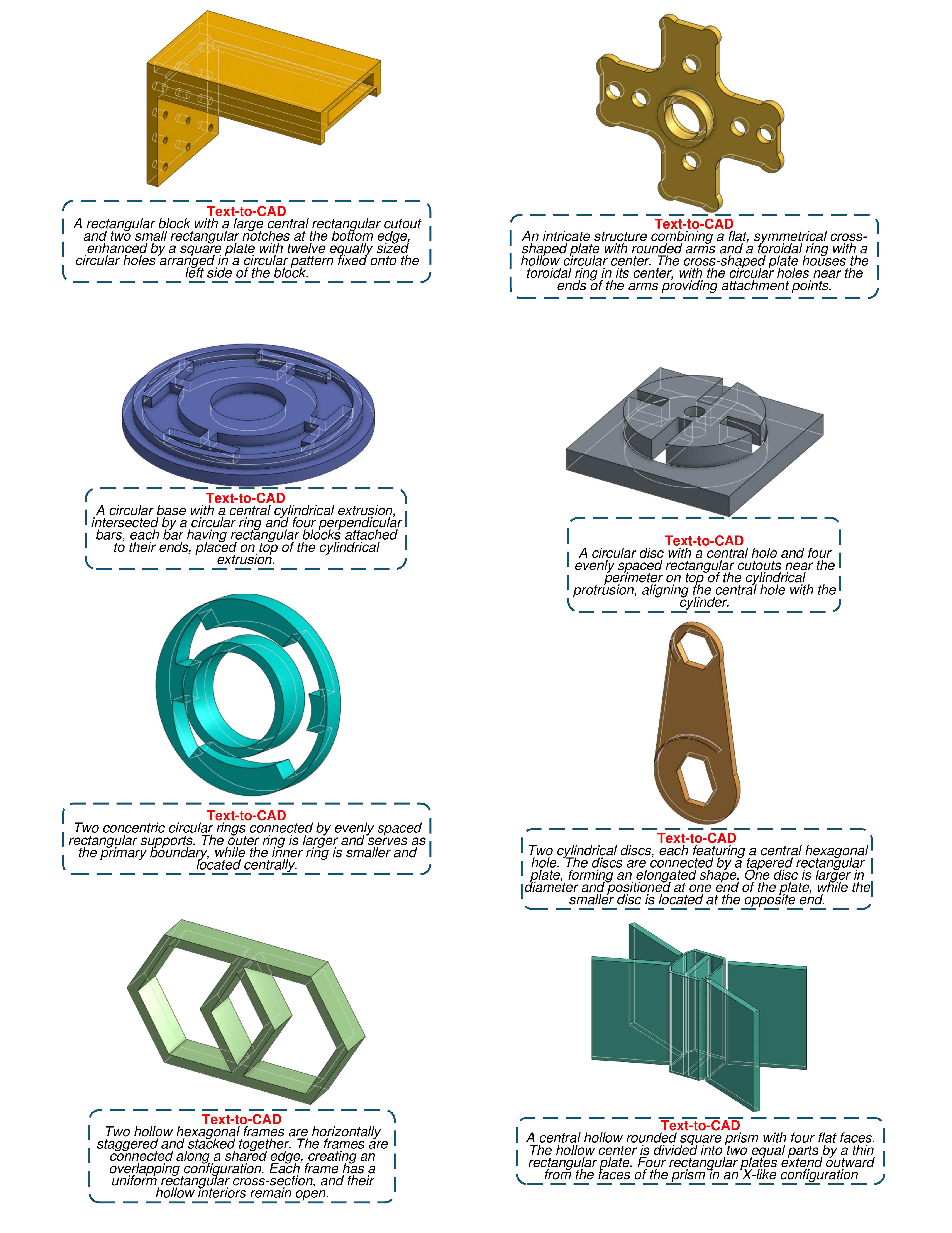}
	\caption{Supplementary results of the Text-to-CAD task generated by CAD-Llama-INS based on text prompts.}
 \label{fig:text2cad_3}
\end{figure*}

\begin{figure*}[h]
    \centering
    \vspace{-10mm}
	\includegraphics[width=\linewidth]{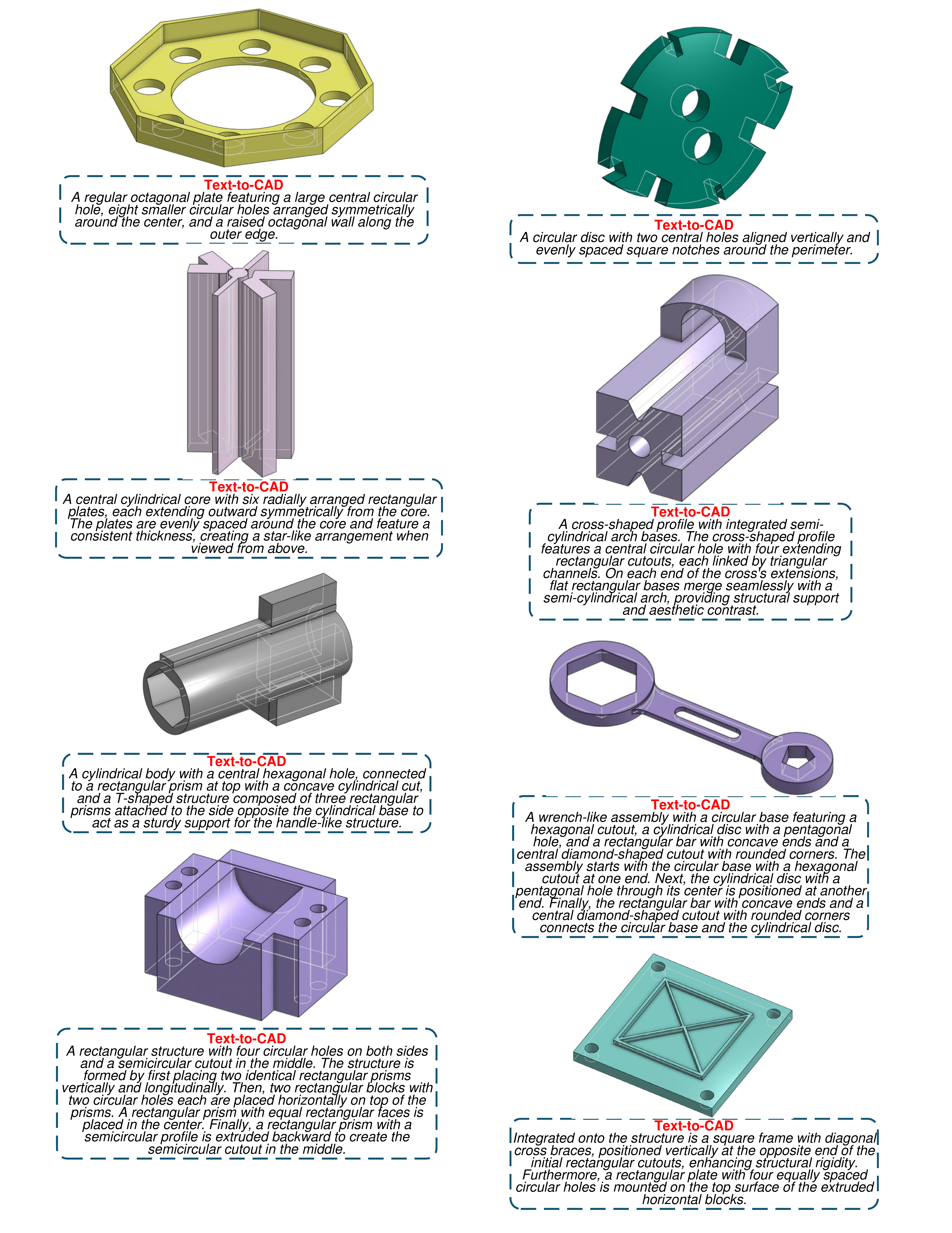}
	\caption{Supplementary results of the Text-to-CAD task generated by CAD-Llama-INS based on text prompts.}
 \label{fig:text2cad_4}
\end{figure*}

\begin{figure*}[h]
    \centering
    \vspace{-10mm}
	\includegraphics[width=\linewidth]{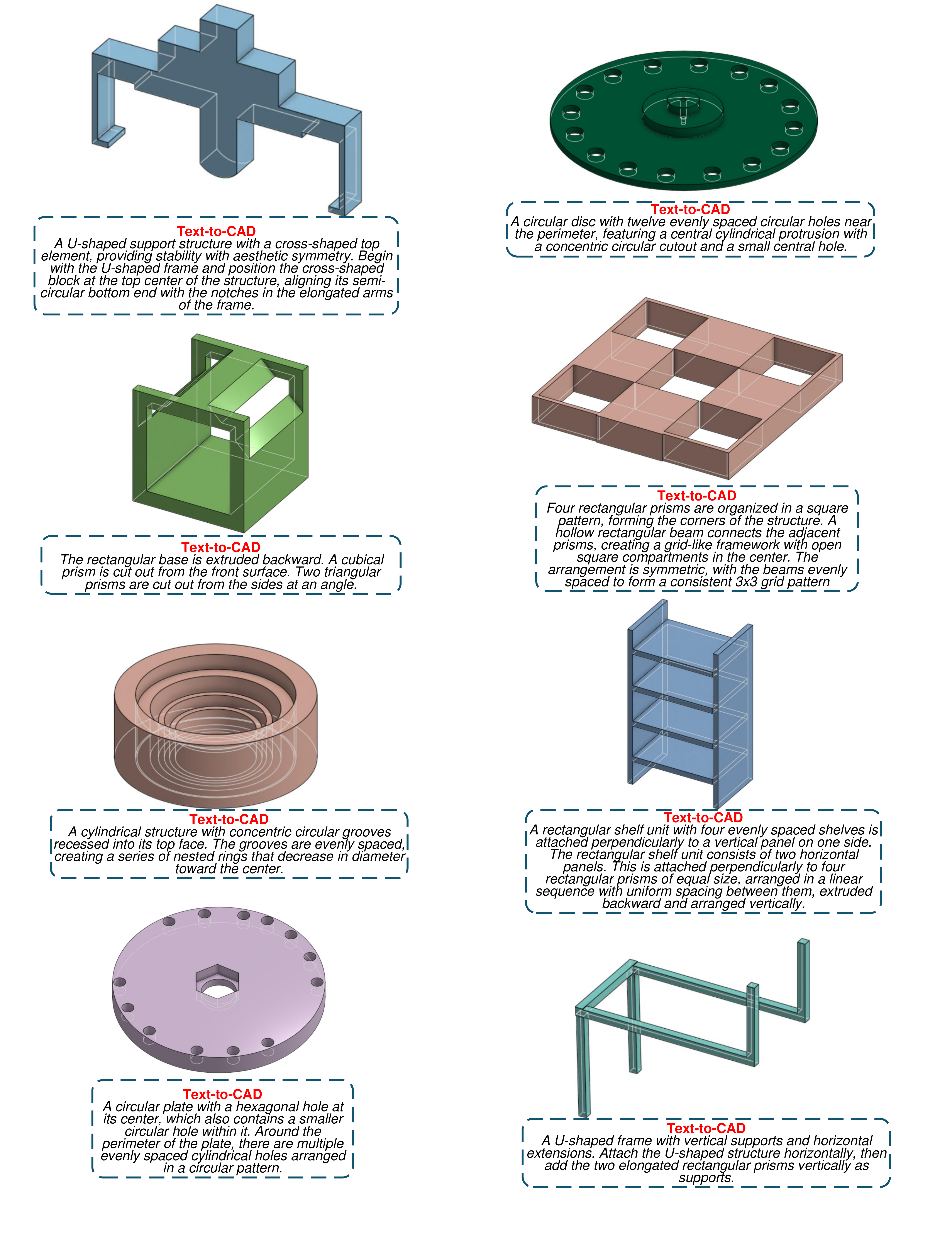}
	\caption{Supplementary results of the Text-to-CAD task generated by CAD-Llama-INS based on text prompts.}
 \label{fig:text2cad_5}
\end{figure*}

\begin{figure*}[h]
    \centering
    \vspace{-10mm}
	\includegraphics[width=\linewidth]{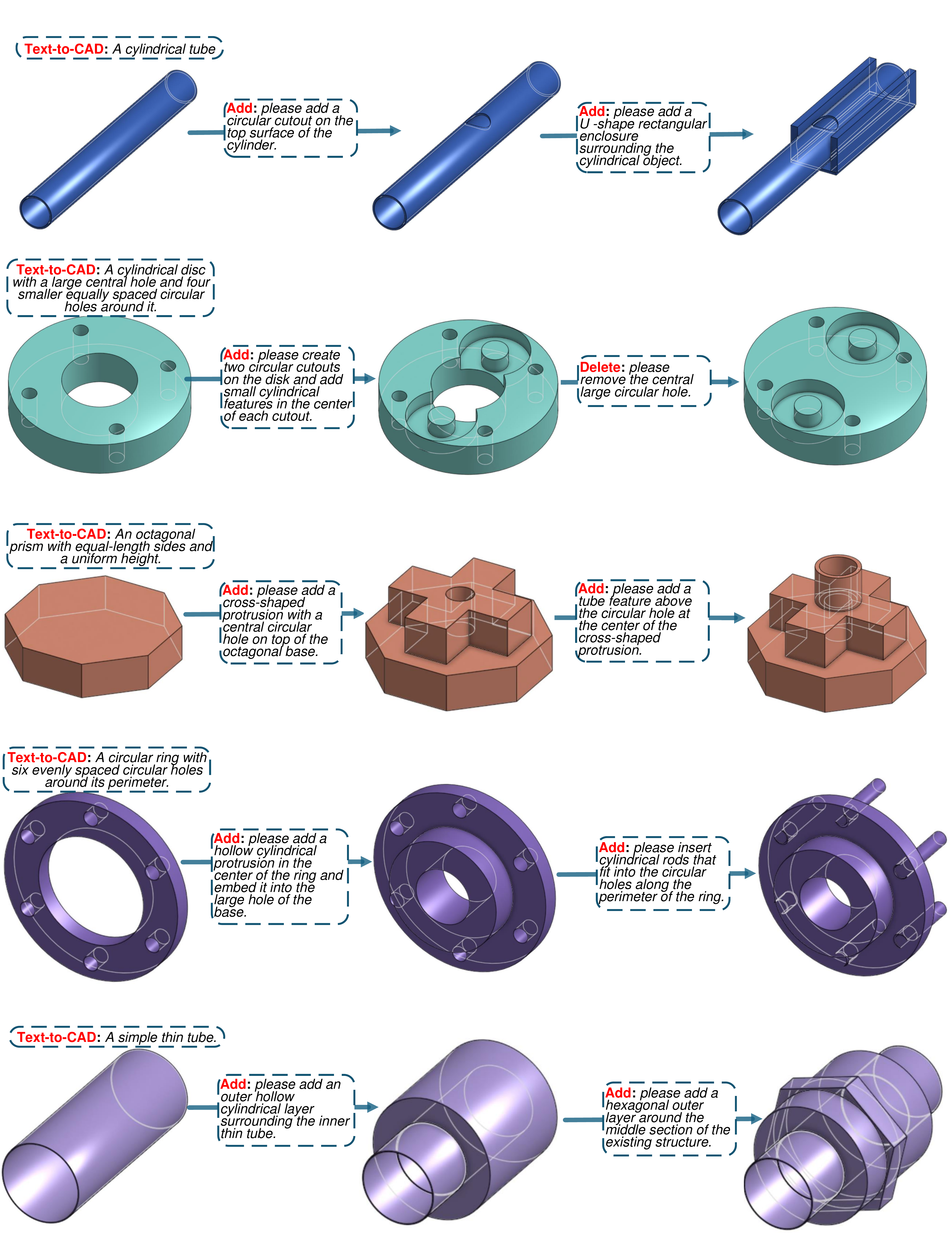}
	\caption{Supplementary working examples of Text-to-CAD, Delete, Add tasks using CAD-Llama-INS.}
 \label{fig:working_example_0}
\end{figure*}

\begin{figure*}[h]
    \centering
    \vspace{-10mm}
	\includegraphics[width=\linewidth]{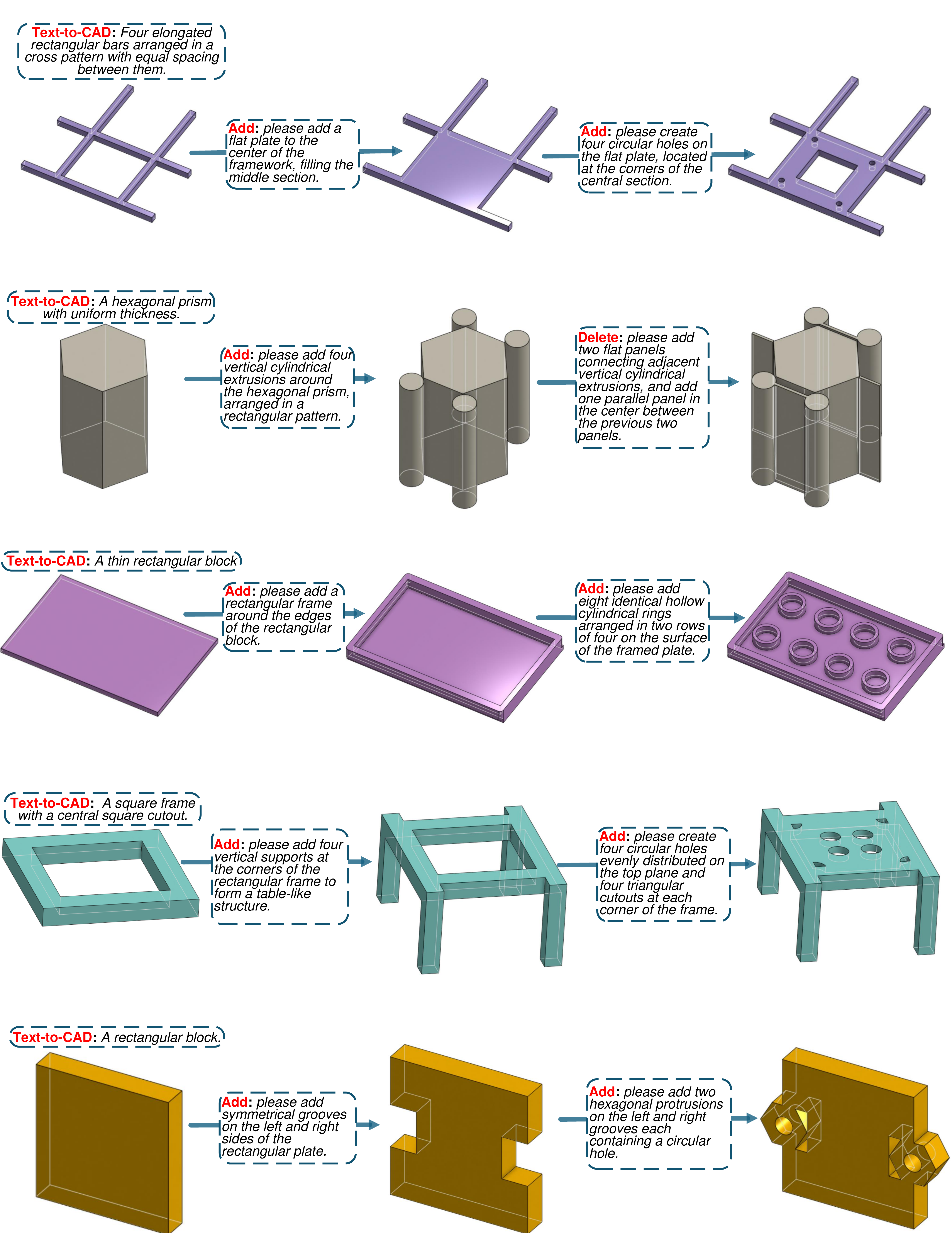}
	\caption{Supplementary working examples of Text-to-CAD, Delete, Add tasks using CAD-Llama-INS.}
 \label{fig:working_example_1}
\end{figure*}

\begin{figure*}[h]
    \centering
    \vspace{-10mm}
	\includegraphics[width=\linewidth]{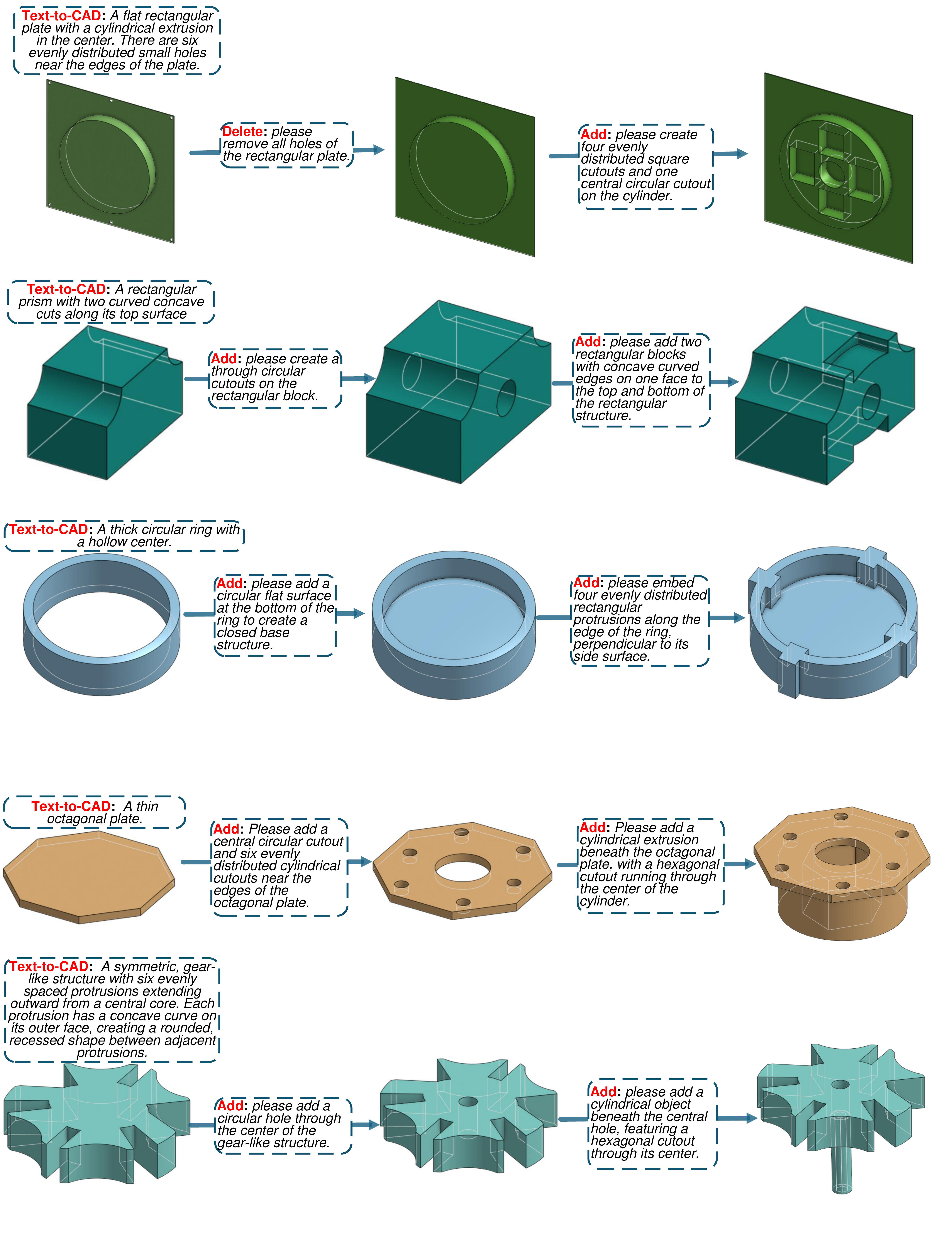}
	\caption{Supplementary working examples of Text-to-CAD, Delete, Add tasks using CAD-Llama-INS.}
 \label{fig:working_example_2}
\end{figure*}

\begin{figure*}[h]
    \centering
    \vspace{-10mm}
	\includegraphics[width=\linewidth]{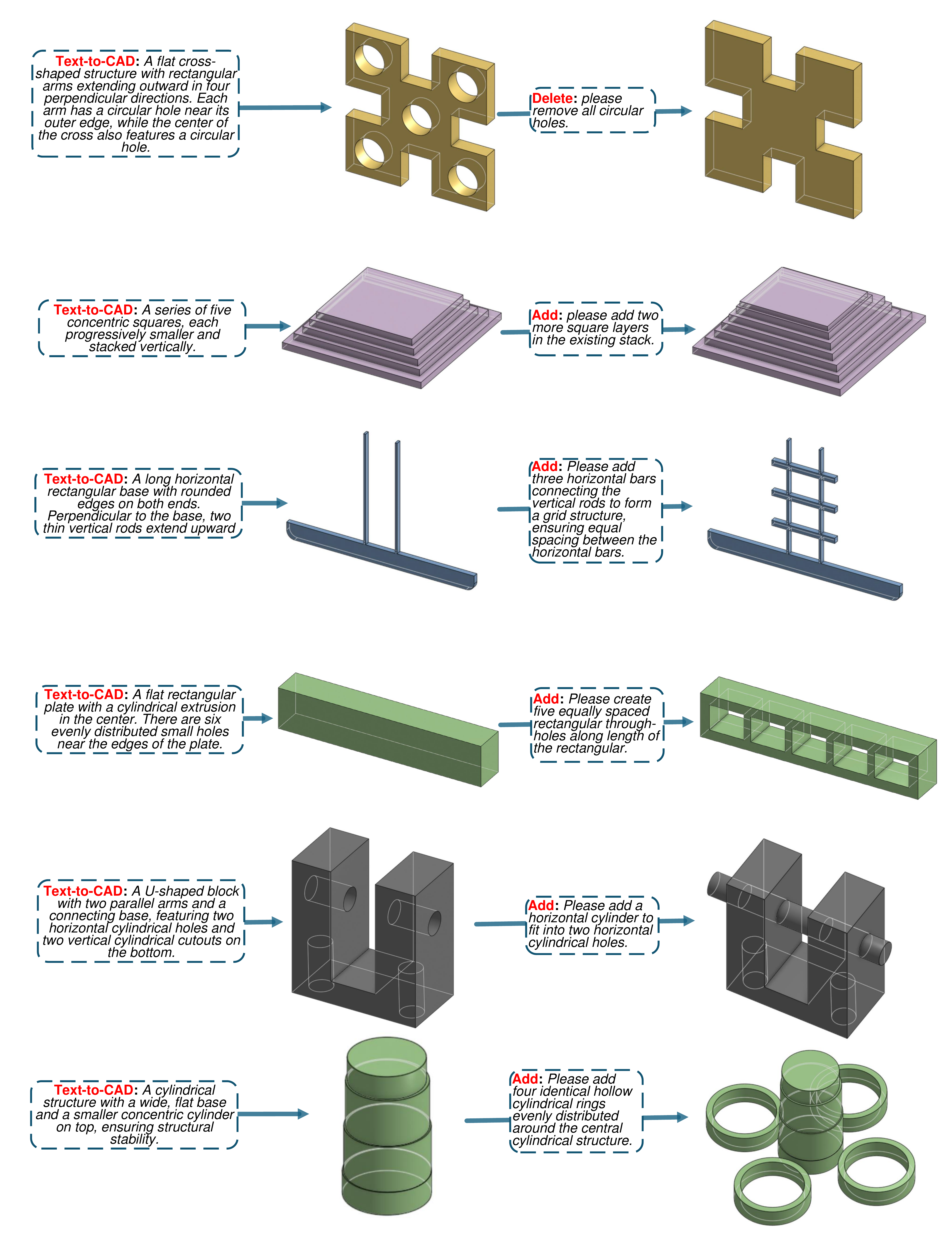}
	\caption{Supplementary working examples of Text-to-CAD, Delete, Add tasks using CAD-Llama-INS.}
 \label{fig:working_example_3}
\end{figure*}

\begin{figure*}[h]
    \centering
    \vspace{-10mm}
	\includegraphics[width=\linewidth]{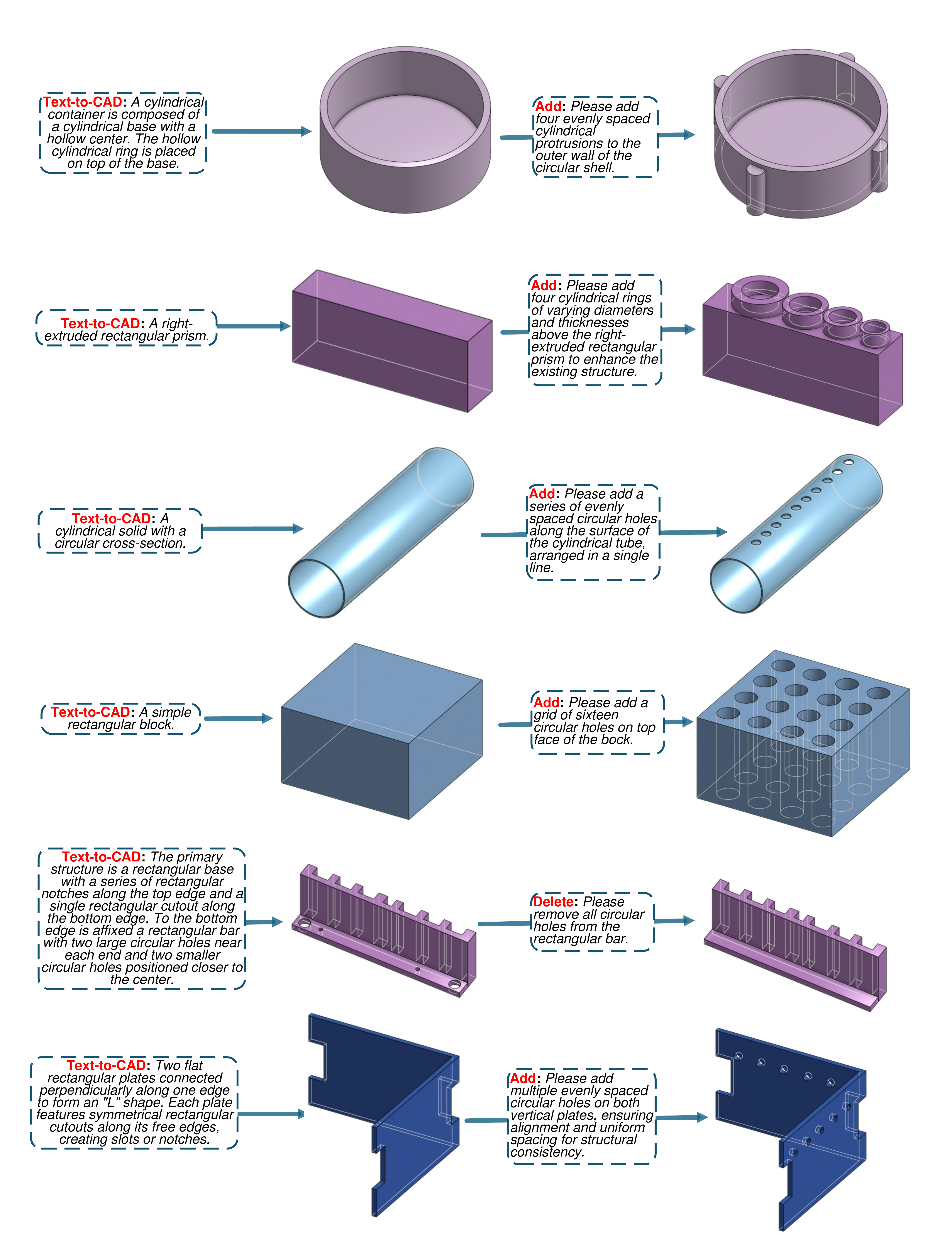}
	\caption{Supplementary working examples of Text-to-CAD, Delete, Add tasks using CAD-Llama-INS.}
 \label{fig:working_example_4}
\end{figure*}

\begin{figure*}[h]
    \centering
    \vspace{-10mm}
	\includegraphics[width=\linewidth]{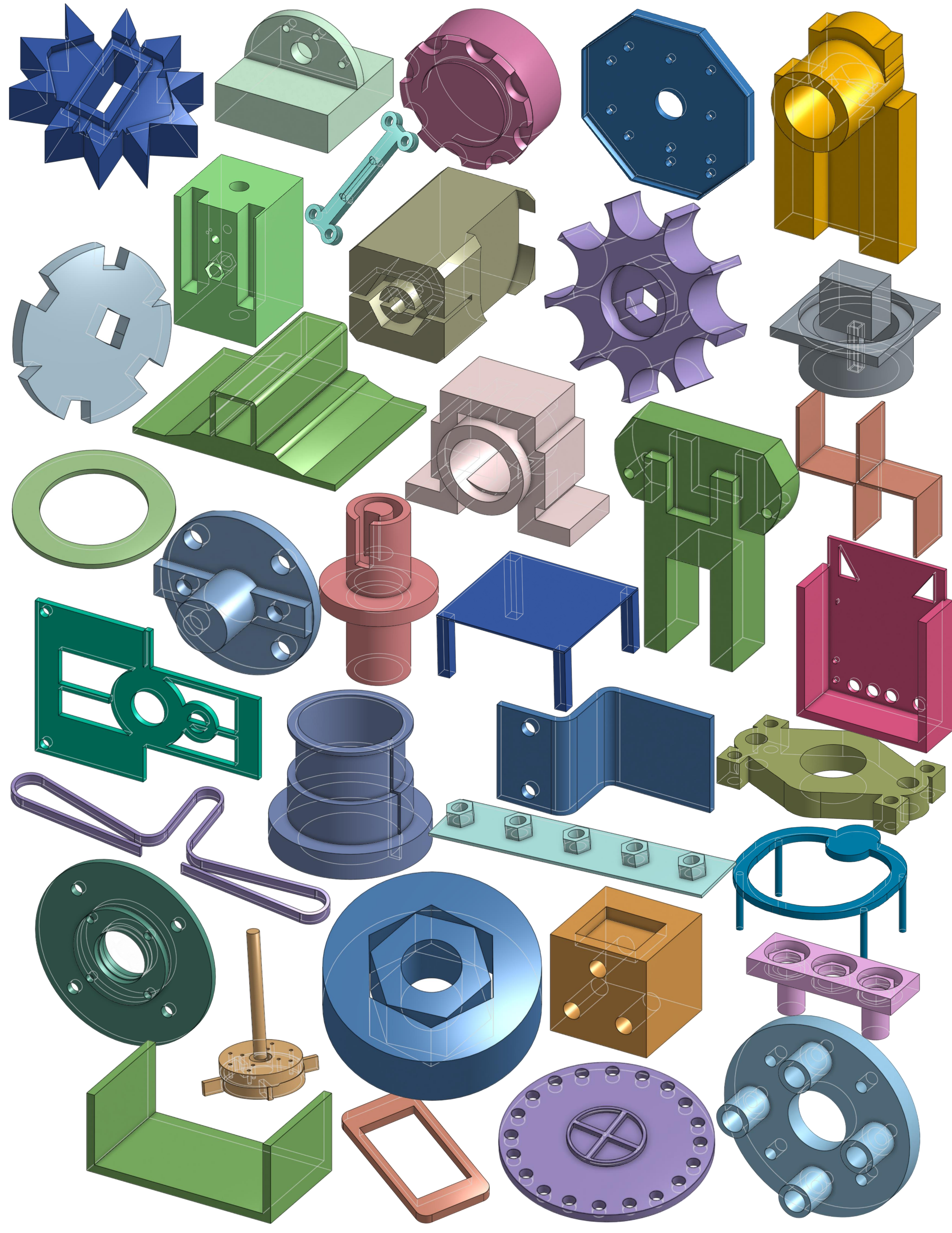}
	\caption{Supplementary results of unconditional generation produced by CAD-Llama.}
 \label{fig:uncond_res}
\end{figure*}

\begin{figure*}[h]
    \centering
	\includegraphics[width=\linewidth]{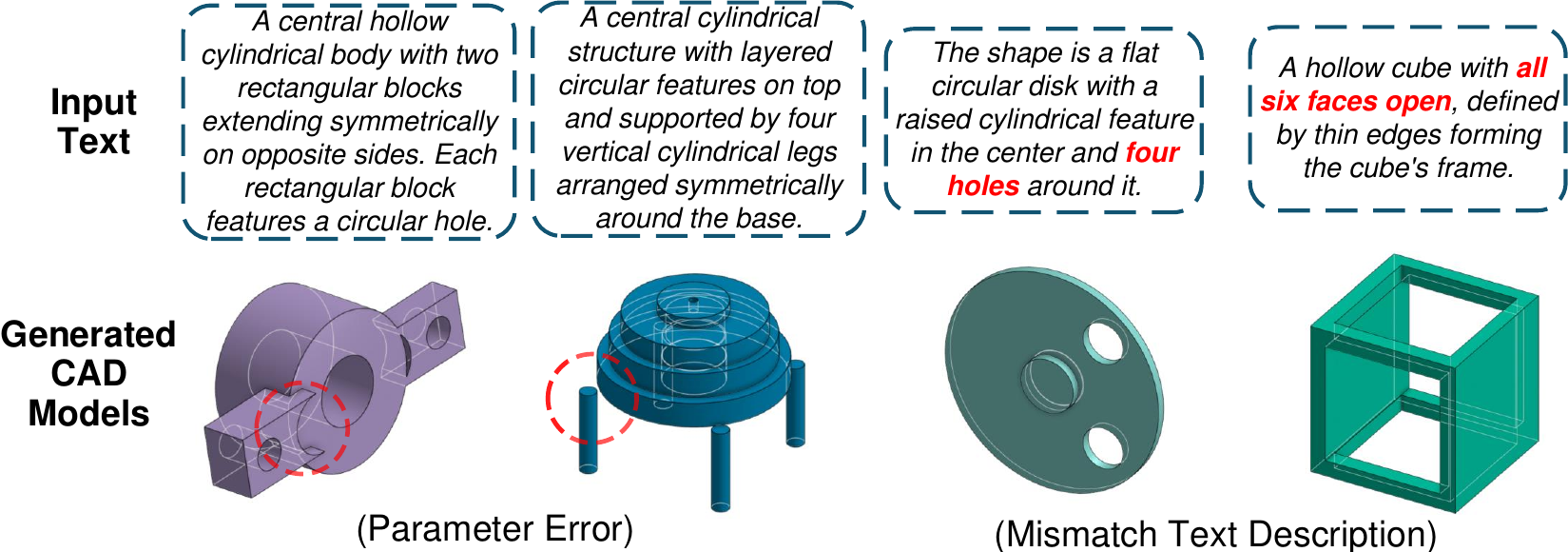}
	\caption{Failure cases for CAD-Llama-INS. We illustrate two types of errors: inaccuracies in parameter settings and misalignment with the text prompts.}
 \label{fig:fail_case}
\end{figure*}

\end{document}